\pdfoutput=1
\documentclass[journal]{IEEEtran}
\usepackage{cite}
\usepackage{amsmath,amssymb,amsfonts}
\usepackage{algorithmic}
\usepackage{graphicx}
\usepackage{textcomp}
\usepackage{tikz}
\usetikzlibrary{shapes.geometric, arrows.meta, positioning, fit, backgrounds, calc, decorations.pathreplacing, patterns, chains}
\usepackage{booktabs}
\usepackage{float}

\newcommand{\history}[1]{}
\newcommand{\doi}[1]{}
\newcommand{\tfootnote}[1]{}
\newcommand{\corresp}[1]{}

\newenvironment{keywords}{\begin{IEEEkeywords}}{\end{IEEEkeywords}}
\newcommand{\EOD}{}
\let\PARstart\IEEEPARstart

\newcommand{\vect}[1]{\mathbf{#1}}
\newcommand{\mat}[1]{\mathbf{#1}}

\def\BibTeX{{\rm B\kern-.05em{\sc i\kern-.025em b}\kern-.08em
    T\kern-.1667em\lower.7ex\hbox{E}\kern-.125emX}}

\begin{document}

\title{\textit{\small Accepted for publication in IEEE Access. DOI: 10.1109/ACCESS.2026.3691980. \copyright\ 2026 IEEE.}\\[0.5em]Stock Market Prediction Using Node Transformer Architecture Integrated with BERT Sentiment Analysis}
\author{\IEEEauthorblockN{Mohammad Al Ridhawi, Mahtab Haj Ali, and Hussein Al Osman}\\
\IEEEauthorblockA{School of Electrical Engineering and Computer Science,\\University of Ottawa, Ottawa, Canada\\e-mail: malri039@uottawa.ca}}

\markboth
{}
{}

\maketitle

\begin{abstract}
Stock market prediction presents considerable challenges for investors, financial institutions, and policymakers operating in complex market environments characterized by noise, non-stationarity, and behavioral dynamics. Traditional forecasting methods, including fundamental analysis and technical indicators, often fail to capture the intricate patterns and cross-sectional dependencies inherent in financial markets. This paper presents an integrated framework combining a node transformer architecture with Bidirectional Encoder Representations from Transformers (BERT)-based sentiment analysis for stock price forecasting. The proposed model represents the stock market as a graph structure where individual stocks form nodes and edges capture relationships including sectoral affiliations, correlated price movements, and supply chain connections. A fine-tuned BERT model extracts sentiment information from social media posts and combines it with quantitative market features through attention-based fusion mechanisms. The node transformer processes historical market data while capturing both temporal evolution and cross-sectional dependencies among stocks. Experiments conducted on 20 S\&P 500 stocks spanning January 1982 to March 2025 demonstrate that the integrated model achieves a mean absolute percentage error (MAPE) of 0.80\% for one-day-ahead predictions, compared to 1.20\% for ARIMA and 1.00\% for LSTM. The inclusion of sentiment analysis reduces prediction error by 10\% overall and 25\% during earnings announcements, while the graph-based architecture contributes an additional 15\% improvement by capturing inter-stock dependencies. Directional accuracy reaches 65\% for one-day forecasts. Statistical validation through paired t-tests confirms the significance of these improvements ($p < 0.05$ for all comparisons). The model maintains lower error during high-volatility periods, achieving MAPE of 1.50\% while baseline models range from 1.60\% to 2.10\%.
\end{abstract}


\section{Introduction}
\label{sec:introduction}

\PARstart{S}{tock} price forecasting represents a fundamental challenge in quantitative finance with substantial implications for portfolio management, risk assessment, and capital allocation \cite{fama1970efficient}. The task of predicting stock market movements is complicated by several interconnected factors: non-stationary price dynamics, high-dimensional feature spaces, complex inter-stock dependencies, and the influence of investor psychology on market behavior.

\subsection{Challenges in Stock Price Forecasting}

A primary challenge in financial forecasting is the presence of noise and irregularities in market data. Microstructure noise, arising from bid-ask bounces and discrete price movements, significantly impacts short-term price observations \cite{zhang2022stock}. Random fluctuations that do not reflect fundamental changes in asset value can obscure meaningful signals, complicating model training and evaluation. The efficient market hypothesis posits that stock prices incorporate all available information, suggesting that consistently outperforming market benchmarks through prediction is theoretically difficult \cite{fama1970efficient}. Empirical evidence indicates partial efficiency, with information incorporated into prices at varying speeds depending on asset liquidity, market conditions, and information type \cite{martin2022efficiency, goldstein2023information}.

Behavioral factors introduce additional complexity. Investor sentiment and cognitive biases lead to market movements not easily explained by quantitative fundamentals alone \cite{kahneman1979prospect}. Fear, greed, and herd behavior drive price fluctuations in patterns that resist traditional modeling approaches. These behavioral dynamics manifest particularly during market stress, when correlations among assets increase and volatility clusters temporally.

The regulatory environment compounds these difficulties. Shifts in financial regulations, monetary policy, and fiscal interventions can trigger sudden market movements that invalidate learned patterns. The 2008 financial crisis and 2020 pandemic response demonstrated how policy actions create discontinuities in price dynamics that historical models may not anticipate.

\subsection{Existing Approaches}

Current approaches to stock price forecasting span multiple methodological traditions. Fundamental analysis evaluates company financial health and market position to estimate intrinsic value \cite{murphy1999technical}. Technical analysis examines historical price and volume patterns to identify trading signals. Machine learning methods, including support vector machines and ensemble methods, apply algorithms to discover patterns in feature-engineered datasets \cite{zhang2022stock}.

Deep learning architectures have demonstrated particular promise in capturing temporal dependencies. Recurrent neural networks, especially Long Short-Term Memory (LSTM) networks, model sequential data through gating mechanisms that address vanishing gradient problems \cite{fischer2018deep}. While effective for short-to-medium range dependencies, LSTMs process sequences step by step, which limits their ability to capture interactions between distant time steps efficiently. The transformer architecture \cite{vaswani2017attention} overcomes this limitation through self-attention mechanisms that relate all positions in a sequence directly, enabling explicit modeling of long-range dependencies without recurrent connections. In parallel, graph neural networks have emerged to address a different shortcoming: the inability of standard sequence models to represent relationships between entities \cite{wu2020comprehensive}. The node transformer architecture bridges these two lines of work by combining attention mechanisms with graph structure, learning contextualized node representations through attention applied over graph neighborhoods \cite{wu2022nodeformer}.

\subsection{Research Motivation and Contributions}

These advances notwithstanding, the literature reveals persistent shortcomings. Few approaches combine graph-based modeling of inter-stock relationships with transformer architectures for temporal modeling. The integration of unstructured textual data, particularly social media sentiment, with structured numerical data remains limited. Performance often deteriorates during volatile market conditions, with accuracy varying substantially between stable and turbulent periods.

This paper presents an integrated framework addressing these gaps through multiple coordinated components:

\begin{enumerate}
\item A node transformer architecture adapted for stock price forecasting that represents the market as a dynamic graph with learnable edge weights. The model captures interdependencies among stocks including sectoral relationships, correlated price movements, and supply chain connections while processing multivariate stock-level inputs.

\item Integration of Bidirectional Encoder Representations from Transformers (BERT)-based sentiment analysis through multimodal fusion, enabling incorporation of qualitative market sentiment alongside quantitative price and volume data. Sentiment information extracted from social media posts is incorporated as additional features for each stock node.

\item An attention-based fusion mechanism that dynamically weights price-based features and sentiment signals based on current market conditions and volatility indicators.

\item A comprehensive dataset comprising historical prices for 20 S\&P 500 companies from January 1982 to March 2025, augmented with sentiment scores derived from social media posts from 2007 onward.

\item Extensive experimental validation across different market conditions, prediction horizons, and volatility regimes, with statistical tests confirming significance of performance improvements.
\end{enumerate}

The remainder of this paper is organized as follows. Section~\ref{sec:literature} reviews related work. Section~\ref{sec:datasets} describes datasets and preprocessing. Section~\ref{sec:methodology} presents the methodology. Section~\ref{sec:experiments} reports experimental results. Section~\ref{sec:discussion} discusses findings and limitations. Section~\ref{sec:conclusion} concludes and outlines future directions.

\section{Literature Review}
\label{sec:literature}

Stock price forecasting has evolved through statistical modeling, machine learning, and deep learning paradigms, each offering distinct trade-offs between interpretability, capacity, and data requirements. The following subsections trace this progression and identify the gaps that motivate the present work.

\subsection{Statistical Methods}

Autoregressive Integrated Moving Average (ARIMA) models remain widely used due to their capacity to capture linear temporal dependencies and interpretable coefficient structures. ARIMA combines autoregressive (AR) and moving average (MA) components after differencing to achieve stationarity. The ARIMA$(p,d,q)$ model is specified as:

\begin{equation}
\label{eq:arima}
\phi(B)(1-B)^d X_t = \theta(B)\epsilon_t
\end{equation}

where $B$ is the backshift operator such that $BX_t = X_{t-1}$, $\phi(B) = 1 - \phi_1 B - \cdots - \phi_p B^p$ is the AR polynomial of order $p$, $\theta(B) = 1 + \theta_1 B + \cdots + \theta_q B^q$ is the MA polynomial of order $q$, $d$ is the differencing order, and $\epsilon_t \sim \mathcal{N}(0, \sigma^2)$ is white noise.

Low and Sakk \cite{low2023comparison} compared ARIMA and LSTM networks for stock prediction across ten exchange-traded funds, finding that ARIMA demonstrates comparable accuracy to LSTM for longer-term predictions. Wahyudi \cite{wahyudi2017arima} applied ARIMA to Indonesian stock prices, demonstrating effectiveness for short-term volatility prediction with model selection via the Akaike Information Criterion (AIC).

Vector Autoregression (VAR) extends univariate ARIMA to multivariate settings, modeling each variable as a linear function of its own lags and lags of other variables. However, ARIMA and VAR are limited in handling nonlinear dynamics, regime shifts, and high-dimensional interactions.

\subsection{Classical Machine Learning}

Support Vector Machines (SVM), k-Nearest Neighbors (KNN), and ensemble methods leverage engineered features for prediction. Nayak et al. \cite{nayak2015naive} developed a hybrid SVM-KNN model for Indian stock market indices, outperforming individual models in mean squared error. Siddique and Panda \cite{siddique2019prediction} combined kernel Principal Component Analysis (PCA) with Support Vector Regression (SVR) and teaching-learning-based optimization, achieving improved accuracy through dimension reduction.

Basak et al. \cite{basak2019predicting} applied XGBoost for long-term stock price forecasting using technical indicators, achieving F-scores ranging from 0.82 to 0.97 for 60-day and 90-day prediction horizons on Apple and Yahoo stocks. While effective for specific tasks, these methods rely heavily on feature engineering and often struggle to generalize across different market regimes.

\subsection{Deep Learning Methods}

\subsubsection{Convolutional Neural Networks}

Convolutional Neural Networks (CNNs) extract hierarchical patterns from structured data through local receptive fields and parameter sharing. Chandar \cite{chandar2022cnn} proposed TI-CNN, converting technical indicators into images using Gramian Angular Fields before CNN processing. Wen et al. \cite{wen2019stock} applied CNNs to S\&P 500 data, exploiting spatial structure in time series representations. CNNs capture local dependencies effectively but are less suited for modeling long-range temporal dynamics.

\subsubsection{Recurrent Neural Networks}

Recurrent Neural Networks (RNNs), particularly LSTM networks and Gated Recurrent Units (GRU), model sequential data through hidden states that propagate information across time steps. LSTM networks incorporate input, forget, and output gates to control information flow, addressing the vanishing gradient problem inherent in standard RNNs.

Di Persio and Honchar \cite{dipersio2017recurrent} compared RNN variants for Google stock prediction, with LSTM achieving 72\% directional accuracy, outperforming standard RNN and GRU architectures. Hossain et al. \cite{hossain2018hybrid} applied cascaded LSTM-GRU to S\&P 500 data spanning 1950-2016, achieving mean squared error of 0.00098. Bidirectional LSTM extends this by processing sequences in both directions. Xu et al. \cite{xu2020stacked} integrated wavelet transform, stacked autoencoder, and BiLSTM for stock forecasting. Liu et al. \cite{liu2022bilstm} employed autoencoder-based feature extraction with BiLSTM for price series prediction.

\subsection{Graph Neural Networks}

Graph Neural Networks (GNNs) capture interdependencies by modeling markets as graphs where nodes represent companies and edges represent relationships such as sectoral ties, price correlations, or supply chain connections. Kipf and Welling \cite{kipf2017semi} introduced the graph convolutional network (GCN) framework, which aggregates features from neighboring nodes through symmetric normalization of the adjacency matrix, providing a scalable approach to semi-supervised learning on graph-structured data. This foundational architecture enabled subsequent applications to financial markets, where inter-stock relationships form natural graph topologies.

Chen et al. \cite{chen2021graph} proposed a graph convolutional feature-based CNN model combining graph convolutional networks with dual CNNs to capture market-level and stock-level features simultaneously. Testing on Chinese stocks demonstrated superior performance compared to non-graph approaches. Wang et al. \cite{wang2022multigraph} introduced a multi-graph convolutional neural network defining both static and dynamic correlation graphs. Testing on 42 Chinese market indices showed average prediction error reduction of 5.11\% compared to LSTM and attention-based baselines. Beyond equity markets, graph-based forecasting methods have been applied to other financial domains. Le et al. \cite{le2025egcn} developed an entropy-based GCN framework for real estate market prediction that segments spatial regions into anomalous and normal clusters before applying separate forecasting models, reducing prediction error across multiple horizons on U.S., U.K., and Australian housing data. A related study by Le et al. \cite{le2025enhancing} combined entropy-based multivariate clustering with sentiment extracted from news headlines and macroeconomic indicators, demonstrating that this segmentation approach improves forecasting accuracy when paired with deep learning architectures such as transformers and LSTMs. These cross-domain results underscore the generality of graph-based representations for capturing complex dependencies in financial data.

\subsection{Transformer Models}

The transformer architecture \cite{vaswani2017attention} introduced a fundamentally different approach to sequence-to-sequence modeling. Self-attention mechanisms compute pairwise interactions between all positions:

\begin{equation}
\label{eq:attention}
\text{Attention}(\mat{Q}, \mat{K}, \mat{V}) = \text{softmax}\left(\frac{\mat{Q}\mat{K}^T}{\sqrt{d_k}}\right)\mat{V}
\end{equation}

where $\mat{Q}$, $\mat{K}$, $\mat{V}$ are query, key, and value matrices, and $d_k$ is the key dimension. The scaling factor $\sqrt{d_k}$ prevents dot products from growing large in magnitude.

Multi-head attention extends this by projecting queries, keys, and values multiple times with different learned projections:

\begin{equation}
\label{eq:multihead}
\text{MultiHead}(\mat{Q}, \mat{K}, \mat{V}) = \text{Concat}(\text{head}_1, \ldots, \text{head}_h)\mat{W}^O
\end{equation}

where $\text{head}_i = \text{Attention}(\mat{Q}\mat{W}_i^Q, \mat{K}\mat{W}_i^K, \mat{V}\mat{W}_i^V)$.

Li and Qian \cite{li2019financial} developed a financial transformer processing multiple timeframes simultaneously, improving sensitivity to regime changes. The node transformer \cite{wu2022nodeformer} extends attention to graph-structured data, learning contextualized node representations through attention over graph neighborhoods.

\subsection{Hybrid Models and Research Gaps}

Hybrid models seek to integrate complementary architectural strengths by pairing spatial feature extractors with sequential learners. Lu et al. \cite{lu2021cnn} proposed a CNN-BiLSTM-attention model that uses convolutional layers to extract local patterns, feeds them into a bidirectional LSTM for temporal modeling, and applies attention to weight the most informative time steps, outperforming standalone approaches on multiple benchmarks. These composite designs show particular promise during volatile periods, where the diversity of learned representations confers robustness that single-paradigm models lack.

More recent work has explored integrating transformer attention with sentiment signals and privacy-preserving training strategies. Nejatbakhsh and Aliasgari \cite{nejatbakhsh2025enhancing} combined LSTM networks with transformer-based attention and FinBERT-derived sentiment in a federated learning framework, achieving an average $R^2$ of 0.91 across ten technology stocks. Their approach demonstrated that sentiment integration improves short-term forecasting accuracy, though the architecture treats each stock independently without modeling inter-stock relationships, and the evaluation was restricted to a single sector. These findings reinforce the value of multimodal fusion while highlighting the continued need for architectures that simultaneously capture cross-asset dependencies.

Despite progress along each of these fronts, critical gaps persist at their intersections. Few approaches combine graph-based inter-stock modeling with transformer temporal processing within a unified architecture. While sentiment integration has been explored alongside recurrent and attention-based models, simultaneous learning of inter-stock graph structure, long-range temporal patterns, and textual sentiment remains largely unaddressed. Performance often deteriorates during volatile conditions, with accuracy varying by 40\% between stable and turbulent periods \cite{low2023comparison}. Computational demands present additional barriers to real-time deployment.

The present work addresses these gaps through integration rather than through the introduction of a fundamentally new learning primitive. Each individual component, namely graph structure learning, transformer-based attention, and BERT sentiment analysis, draws on established techniques. The contribution lies in the architectural design required to combine them within a single end-to-end framework. The node transformer must compute attention simultaneously across temporal positions and graph neighborhoods, which requires adapting the standard self-attention mechanism to operate over a three-dimensional input tensor rather than a flat sequence. The adaptive gating mechanism that fuses continuous sentiment scores with numerical price features handles temporal sparsity and missing data without degrading the price-based pathway, a challenge that simple feature concatenation does not address. The jointly learned edge weights provide an interpretable structural layer connecting model internals to domain knowledge about market relationships, offering a qualitative advantage that pure performance comparisons cannot capture. This integrative design philosophy follows a productive tradition in deep learning, where combining known components in carefully engineered architectures has yielded substantial advances over any single technique deployed in isolation.

\section{Datasets and Preprocessing}
\label{sec:datasets}

\subsection{Financial Market Dataset}
\label{subsec:fmd}

The Financial Market Dataset (FMD) integrates historical stock market data with engineered technical indicators. The dataset spans January 1, 1982 to March 31, 2025, covering twenty companies from the S\&P 500 index selected for sectoral diversity, market capitalization variation, and data availability.

The selected companies span multiple sectors: technology (Apple, Microsoft, Salesforce, Netflix), financial services (JPMorgan Chase, Visa), healthcare (Johnson \& Johnson, UnitedHealth Group, Pfizer), retail (Walmart, Home Depot, McDonald's), energy (ExxonMobil, Chevron), consumer goods (Procter \& Gamble, Coca-Cola, Nike), telecommunications (Verizon), and industrials (Boeing, Caterpillar). For companies with initial public offerings after 1982, data collection begins at the first available trading date. For example, Salesforce data begins in June 2004 following its IPO. The effective analysis period for each stock corresponds to its available trading history within the 1982-2025 window. This cross-sectional design ensures the dataset reflects behavior across multiple industries and economic cycles. The 20-stock universe was chosen deliberately to balance temporal depth with cross-sectional coverage: a smaller, well-established set of equities allows for training on over four decades of data with complete and reliable records, which would be difficult to achieve with a much larger or historically reconstructed universe where data quality varies substantially. This design is best understood as a proof-of-concept demonstrating the architecture's capacity to learn inter-stock dependencies and integrate sentiment signals, rather than a comprehensive market simulation. The selection of current index constituents does introduce survivorship bias, which is examined in detail in Section~\ref{sec:discussion}.

For each company, daily Open, High, Low, Close, and Volume (OHLCV) trading variables, along with the adjusted closing price, were collected: opening price ($O_t$), highest price ($H_t$), lowest price ($L_t$), closing price ($C_t$), adjusted closing price ($C_t^{\text{adj}}$), and trading volume ($V_t$). Technical indicators were computed following established methodologies \cite{xu2020stacked, chandar2022cnn}:

\textbf{Simple Moving Average (SMA):} The SMA smooths short-term price fluctuations to reveal underlying trends. Computing it at multiple window lengths ($n \in \{5, 10, 20\}$ days) allows the model to distinguish short-term momentum from longer-term directional shifts:

\begin{equation}
\label{eq:sma}
\text{SMA}_n(t) = \frac{1}{n}\sum_{i=0}^{n-1} C_{t-i}
\end{equation}

\textbf{Exponential Moving Average (EMA):} Unlike the SMA, the EMA assigns greater weight to recent observations through an exponential decay controlled by smoothing factor $\alpha = 2/(n+1)$. This makes it more responsive to recent price changes, which is valuable for detecting early trend reversals:

\begin{equation}
\label{eq:ema}
\text{EMA}_n(t) = \alpha \cdot C_t + (1-\alpha) \cdot \text{EMA}_n(t-1)
\end{equation}

\textbf{Relative Strength Index (RSI):} The RSI is a bounded momentum oscillator that measures the speed and magnitude of recent price changes on a scale from 0 to 100. Values above 70 typically indicate overbought conditions, while values below 30 suggest oversold conditions. It is computed using a 14-day window:

\begin{equation}
\label{eq:rsi}
\text{RSI}(t) = 100 - \frac{100}{1 + \frac{\overline{G}_{14}(t)}{\overline{L}_{14}(t)}}
\end{equation}

where $\overline{G}_{14}(t)$ and $\overline{L}_{14}(t)$ are the average gains and average losses over the preceding 14 trading days.

\textbf{Moving Average Convergence Divergence (MACD):} The MACD captures changes in trend strength and direction by measuring the difference between a short-term (12-day) and a long-term (26-day) EMA. A positive MACD indicates upward momentum, while a negative value signals downward pressure. Crossovers between the MACD and its 9-day signal line are commonly used as trade triggers:

\begin{equation}
\label{eq:macd}
\text{MACD}(t) = \text{EMA}_{12}(t) - \text{EMA}_{26}(t)
\end{equation}

with 9-day signal line $\text{Signal}(t) = \text{EMA}_9(\text{MACD}(t))$.

\textbf{Daily Returns and Log Returns:} Daily returns quantify the relative price change between consecutive trading days and serve as the primary measure of stock performance. Log returns are included alongside arithmetic returns because they are additive over time and more closely approximate a normal distribution, properties that benefit gradient-based optimization:

\begin{equation}
\label{eq:returns}
r_t = \frac{C_t - C_{t-1}}{C_{t-1}}, \quad r_t^{\log} = \ln\left(\frac{C_t}{C_{t-1}}\right)
\end{equation}

\textbf{Rolling Volatility:} Rolling volatility measures the dispersion of returns over a trailing 20-day window, providing a real-time estimate of price uncertainty. Higher volatility indicates greater risk and wider expected price ranges, which is critical for the model's gating mechanism to adjust feature emphasis across market regimes:

\begin{equation}
\label{eq:volatility}
\sigma_t = \sqrt{\frac{1}{19}\sum_{i=0}^{19}(r_{t-i} - \bar{r}_t)^2}
\end{equation}

Missing values were addressed through a temporally aware imputation strategy designed to prevent information leakage. During training data preparation, short gaps (1-2 trading days) used linear interpolation:

\begin{equation}
\label{eq:interpolation}
C_t = C_{t-k} + \frac{C_{t+m} - C_{t-k}}{k + m} \cdot k
\end{equation}

where $k$ and $m$ are distances to nearest observed values. For validation and test data, only forward-filling from the most recent observed value was applied to ensure no future information leaked into predictions: $C_t = C_{t-k}$ where $k$ is the distance to the most recent observation. Longer gaps in training data also employed forward filling. Each observation contains 6 raw variables and 11 derived indicators.

Feature normalization used an expanding window z-score approach to prevent look-ahead bias. For each feature $x_{i,t}$, the normalized value was computed as:

\begin{equation}
\label{eq:zscore}
\tilde{x}_{i,t} = \frac{x_{i,t} - \mu_{1:t}}{\sigma_{1:t}}
\end{equation}

where $\mu_{1:t}$ and $\sigma_{1:t}$ are the mean and standard deviation computed over all available data from the start of the training period up to time $t$. For test data, normalization statistics were fixed using the full training set, ensuring no information from the test period influenced the standardization.

\begin{figure}[ht]
\centering
\begin{tikzpicture}[
    node distance=0.5cm and 0.8cm,
    box/.style={rectangle, draw, minimum width=1.4cm, minimum height=0.5cm, align=center, font=\tiny},
    arrow/.style={-{Stealth[scale=0.5]}, thick}
]
\node[box, fill=blue!15] (raw) {Raw OHLCV\\Data};
\node[box, fill=green!15, below=0.8cm of raw, xshift=-2.5cm] (sma) {SMA\\5, 10, 20};
\node[box, fill=green!15, below=0.8cm of raw, xshift=-0.8cm] (ema) {EMA\\5, 10, 20};
\node[box, fill=green!15, below=0.8cm of raw, xshift=0.8cm] (momentum) {RSI, MACD\\Returns};
\node[box, fill=yellow!20, below=0.8cm of raw, xshift=2.5cm] (vol) {Rolling\\Volatility};
\node[box, fill=orange!20, below=1.0cm of ema, xshift=0.8cm] (norm) {Z-Score\\Normalization};
\node[box, fill=red!15, below=0.5cm of norm] (features) {Feature Vector\\$\vect{x}_{i,t} \in \mathbb{R}^{17}$};
\draw[arrow] (raw) -- (sma);
\draw[arrow] (raw) -- (ema);
\draw[arrow] (raw) -- (momentum);
\draw[arrow] (raw) -- (vol);
\draw[arrow] (sma) -- (norm);
\draw[arrow] (ema) -- (norm);
\draw[arrow] (momentum) -- (norm);
\draw[arrow] (vol) -- (norm);
\draw[arrow] (norm) -- (features);
\end{tikzpicture}
\caption{Feature engineering pipeline. Raw OHLCV data is processed through multiple technical indicator computations, normalized using z-score standardization, and concatenated into the final feature vector.}
\label{fig:feature_pipeline}
\end{figure}
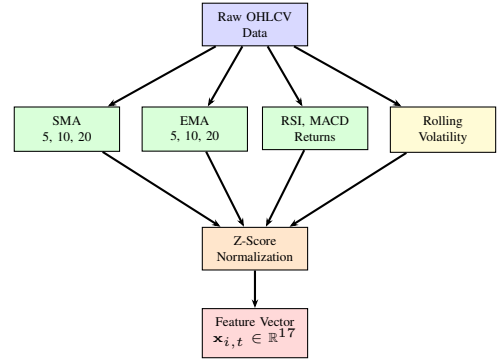

Fig.~\ref{fig:feature_pipeline} illustrates the feature engineering pipeline.

\subsection{Sentiment Datasets}
\label{subsec:sentiment}

The sentiment component relies on two complementary datasets that together capture both benchmark-labeled sentiment and large-scale real-world social media activity. The first provides expert-annotated ground truth for training the BERT classifier, while the second supplies the raw social media text processed at inference time.

The Market Sentiment Evaluation (MSE) dataset is a publicly available corpus containing finance-related messages from social media platforms \cite{cortis2017semeval}. Each message has been annotated by financial experts with sentiment scores $s \in [-1, +1]$, where $-1$ indicates strongly negative, $0$ neutral, and $+1$ strongly positive sentiment. The dataset includes source platform identifiers, message IDs, relevant cashtags, and annotated sentiment spans. Because it contains human-labeled ground truth, the MSE dataset serves as the primary training and validation resource for fine-tuning the BERT sentiment module.

The Comprehensive Stock Sentiment (CSS) dataset was constructed using the X (formerly Twitter) API through systematic searches for the cashtag corresponding to each of the twenty stocks listed in Table~\ref{tab:search_tags}. Since the platform was founded in 2006, the CSS dataset covers January 2007 to March 2025, yielding approximately 4.2 million posts across all stocks.

\begin{table}[!ht]
\centering
\caption{Search tags used for CSS dataset collection.}
\label{tab:search_tags}
\begin{tabular}{llll}
\toprule
\textbf{Cashtag} & \textbf{Company} & \textbf{Cashtag} & \textbf{Company} \\
\midrule
\$AAPL & Apple & \$KO & Coca-Cola \\
\$MSFT & Microsoft & \$CVX & Chevron \\
\$JPM & JPMorgan Chase & \$HD & Home Depot \\
\$JNJ & Johnson \& Johnson & \$VZ & Verizon \\
\$WMT & Walmart & \$BA & Boeing \\
\$XOM & ExxonMobil & \$MCD & McDonald's \\
\$PG & Procter \& Gamble & \$NFLX & Netflix \\
\$V & Visa & \$CAT & Caterpillar \\
\$UNH & UnitedHealth Group & \$PFE & Pfizer \\
\$NKE & Nike & \$CRM & Salesforce \\
\bottomrule
\end{tabular}
\end{table}

Once the BERT model is fine-tuned on the MSE labels, it is applied to the CSS corpus to generate daily sentiment scores for each stock. These scores are then aggregated at multiple time scales and fed into the node transformer as additional input features.

Both datasets underwent identical preprocessing before entering the sentiment pipeline. This included removal of HTML encodings and special characters, anonymization of user identifiers, removal of retweet markers, URL standardization with placeholder tags, de-elongation of exaggerated words (e.g., ``victoryyyyy'' $\rightarrow$ ``victory''), spacing correction for ticker symbols (e.g., ``\$ AAPL'' $\rightarrow$ ``\$AAPL''), and ordinal expression normalization.

\subsection{Dataset Partitioning}
\label{subsec:partitioning}

Temporal splits were applied to prevent look-ahead bias, following the standard practice in financial time series forecasting. Because stock prices exhibit serial dependence and evolving distributional properties, random or stratified partitioning schemes such as k-fold cross-validation would allow future observations to inform predictions of earlier time steps, producing artificially inflated accuracy estimates that do not reflect realistic forecasting conditions. In a forecasting context, the model must never train on data that postdates the prediction target; violating this constraint renders performance metrics unreliable regardless of their magnitude. Chronological partitioning ensures that the model is evaluated exclusively on data that postdates the training window, mirroring the information constraints faced by a real-time forecaster.

The training set comprises January 1, 1982 to December 31, 2010 (70\% of temporal range). The validation set spans January 1, 2011 to December 31, 2016 (15\%), serving as both a hyperparameter tuning period and a temporal buffer that separates training from test data by six calendar years. The test set covers January 1, 2017 to March 31, 2025 (15\%), deliberately spanning multiple distinct market regimes including the pre-pandemic bull market (2017--2019), the COVID-19 crash and recovery (2020--2021), and the inflationary tightening period (2022--2024). The inclusion of the COVID-19 pandemic in the test set rather than in training provides a rigorous assessment of the model's ability to generalize to market conditions qualitatively different from those observed during training. The volatility regime analysis in Section~\ref{subsec:results} confirms that the model maintains reasonable accuracy across these diverse conditions, with MAPE rising from 0.70\% in low-volatility periods to 1.50\% in high-volatility periods, a degradation consistent with the inherent difficulty of forecasting during market stress rather than a failure of generalization.

The full 1982-2025 training window is retained because the price-based components of the model, including the node transformer, graph structure learning, and temporal encoding, benefit substantially from exposure to diverse market regimes spanning over four decades. Restricting training to the post-2007 period would sacrifice the 1982-2006 data that contains the 1987 crash, the dot-com bubble, and the 2008 financial crisis, all of which are critical for learning robust representations of market dynamics.

For sentiment data, partitioning aligns with financial splits for the overlapping period. Since X (formerly Twitter) data begins in 2007, the sentiment training period covers January 2007 to December 2010. For pre-2007 financial data, sentiment features are set to neutral (zero). This zero-padding is handled by the adaptive gating mechanism (Section~\ref{subsec:integration}), which learns to down-weight the sentiment channel when its signal is absent, effectively allowing the model to operate in a price-only mode for the pre-2007 training window and transition to multimodal operation when sentiment data becomes available. The sentiment integration component is primarily evaluated on the test set (2017-2025), where sentiment data is abundant and the model can fully leverage both data streams.

The MSE dataset used for BERT fine-tuning is partitioned separately into 70\% training, 15\% validation, and 15\% test, stratified by sentiment class to preserve label distribution. The BERT model is fine-tuned and evaluated on this split independently before its outputs are integrated into the full forecasting pipeline (see Section~\ref{subsec:bert_eval}). Table~\ref{tab:dataset_summary} summarizes the key characteristics of the financial and sentiment datasets.

\begin{table}[H]
\centering
\caption{Dataset summary statistics.}
\label{tab:dataset_summary}
\begin{tabular}{lcc}
\toprule
\textbf{Characteristic} & \textbf{FMD} & \textbf{CSS} \\
\midrule
Time period & 1982-2025 & 2007-2025 \\
Number of stocks & 20 & 20 \\
Trading days & $\approx$10,300 & $\approx$4,100 \\
Features per stock-day & 17 & 3 \\
Total observations & $\approx$206,000 & $\approx$4.2M posts \\
\bottomrule
\end{tabular}
\end{table}

\section{Methodology}
\label{sec:methodology}

The proposed framework integrates a node transformer architecture with BERT-based sentiment analysis, combining quantitative financial indicators with qualitative textual signals. The design follows a modular structure in which the graph-based transformer and the sentiment pipeline operate as independent components, coupled through an attention-based fusion layer that dynamically adjusts the relative weighting of each stream based on current market conditions.

\subsection{System Architecture Overview}
\label{subsec:overview}

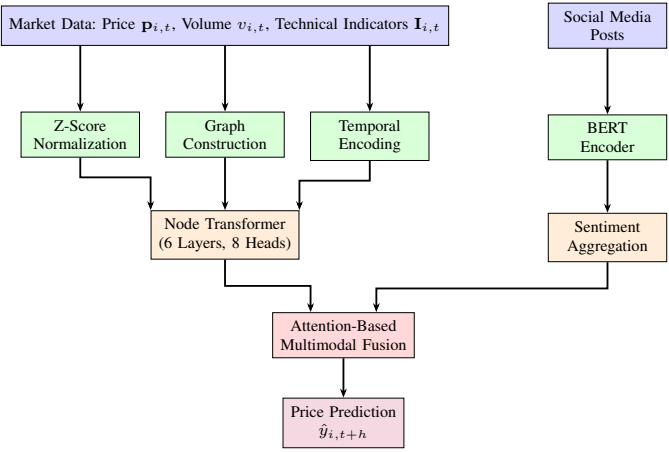
\begin{figure}[!ht]
\centering
\resizebox{\columnwidth}{!}{%
\begin{tikzpicture}[
    node distance=0.5cm and 0.6cm,
    box/.style={rectangle, draw, minimum width=1.8cm, minimum height=0.6cm, align=center, font=\scriptsize},
    bluebox/.style={box, fill=blue!15},
    greenbox/.style={box, fill=green!15},
    orangebox/.style={box, fill=orange!15},
    purplebox/.style={box, fill=purple!15},
    redbox/.style={box, fill=red!15},
    arrow/.style={-{Stealth[scale=0.6]}, thick}
]

\node[bluebox, minimum width=5.5cm] (market) {Market Data: Price $\vect{p}_{i,t}$, Volume $v_{i,t}$, Technical Indicators $\vect{I}_{i,t}$};
\node[bluebox, right=1.5cm of market] (text) {Social Media\\Posts};

\node[greenbox, below=1.0cm of market, xshift=-2.2cm] (norm) {Z-Score\\Normalization};
\node[greenbox, below=1.0cm of market] (graph) {Graph\\Construction};
\node[greenbox, below=1.0cm of market, xshift=2.2cm] (encode) {Temporal\\Encoding};
\node[greenbox, below=1.0cm of text] (bert) {BERT\\Encoder};

\node[orangebox, below=0.8cm of graph] (nodeformer) {Node Transformer\\(6 Layers, 8 Heads)};
\node[orangebox, below=0.8cm of bert] (sentiment) {Sentiment\\Aggregation};

\node[redbox, below=0.8cm of nodeformer, xshift=1.8cm] (fusion) {Attention-Based\\Multimodal Fusion};

\node[purplebox, below=0.6cm of fusion] (pred) {Price Prediction\\$\hat{y}_{i,t+h}$};

\draw[arrow] (market.south) ++(-2.2cm,0) -- (norm.north);
\draw[arrow] (market.south) -- (graph.north);
\draw[arrow] (market.south) ++(2.2cm,0) -- (encode.north);
\draw[arrow] (text) -- (bert);

\draw[arrow] (norm.south) -- ++(0,-0.3cm) -| (nodeformer.north west);
\draw[arrow] (graph) -- (nodeformer);
\draw[arrow] (encode.south) -- ++(0,-0.3cm) -| (nodeformer.north east);
\draw[arrow] (bert) -- (sentiment);

\draw[arrow] (nodeformer.south) -- ++(0,-0.4cm) -| ([xshift=-0.5cm]fusion.north);
\draw[arrow] (sentiment.south) -- ++(0,-0.4cm) -| ([xshift=0.5cm]fusion.north);
\draw[arrow] (fusion) -- (pred);

\end{tikzpicture}%
}
\caption{System architecture. Price data, volume, and technical indicators are jointly processed through normalization, graph construction, and temporal encoding before the node transformer. Social media posts are processed through BERT and sentiment aggregation. Both streams combine through attention-based multimodal fusion.}
\label{fig:architecture}
\end{figure}

Fig.~\ref{fig:architecture} presents the overall system architecture. The framework processes historical market data and sentiment information through two parallel pathways. The quantitative branch normalizes all market features (price, volume, and technical indicators), constructs the graph structure, applies temporal encoding, and feeds the processed representations into the node transformer. The qualitative branch passes social media text through the BERT encoder and aggregates sentiment scores at multiple time scales. The two streams converge at an attention-based fusion layer that produces the final price prediction.

\subsection{Graph Representation of the Stock Market}
\label{subsec:graph}

The stock market is represented as a graph $\mathcal{G} = (\mathcal{V}, \mathcal{E}, \mat{E})$ where $\mathcal{V} = \{1, \ldots, N\}$ is the set of $N = 20$ stock nodes, $\mathcal{E} \subseteq \mathcal{V} \times \mathcal{V}$ is the edge set (fully connected), and $\mat{E} \in \mathbb{R}^{N \times N}$ is the edge weight matrix. The choice of $N = 20$ nodes balances computational tractability with sufficient cross-sectional coverage to capture major sector interactions. While larger graphs might better represent full market topology, the current design enables deep temporal analysis (252-day input sequences) across diverse sectors. The fully-connected structure with learnable edge weights ensures all pairwise relationships can be discovered during training. Ablation results (Section~\ref{subsec:results}) confirm that the graph structure contributes meaningful predictive improvement over models without explicit inter-stock connections.

Edge weights are initialized based on sector relationships and historical price correlations:

\begin{equation}
\label{eq:edge_init}
e_{ij}^{(0)} = \alpha \cdot \delta_{\text{sector}}(i,j) + (1-\alpha) \cdot \max(0, \rho_{ij})
\end{equation}

where $\delta_{\text{sector}}(i,j) = 1$ if stocks $i$ and $j$ share the same Global Industry Classification Standard (GICS) sector classification and $0$ otherwise, $\rho_{ij}$ is the Pearson correlation coefficient between daily returns computed over the training period, $\max(0, \cdot)$ ensures non-negative weights, and $\alpha = 0.5$ balances sector and correlation information.

During training, edge weights are refined iteratively:

\begin{equation}
\label{eq:edge}
e_{ij}^{(\ell+1)} = \sigma\left(\vect{w}_e^T [\vect{h}_i^{(\ell)} \| \vect{h}_j^{(\ell)}] + b_e\right)
\end{equation}

where $\sigma$ is the sigmoid function, $\vect{w}_e \in \mathbb{R}^{2d}$ is a learnable weight vector, $\vect{h}_i^{(\ell)}, \vect{h}_j^{(\ell)} \in \mathbb{R}^d$ are node representations at layer $\ell$, $\|$ denotes concatenation, and $b_e$ is a learnable bias. Edges are undirected: $e_{ij} = e_{ji}$.

At each training step, the model receives a batch of temporal windows. For a single sample, the input is a three-dimensional tensor $\mat{X} \in \mathbb{R}^{T \times N \times d_{\text{in}}}$, where $T = 252$ is the sequence length (trading days), $N = 20$ is the number of stock nodes, and $d_{\text{in}} = 17$ is the per-stock feature dimension after concatenating raw OHLCV data with technical indicators. The edge weight matrix $\mat{E} \in \mathbb{R}^{N \times N}$ is initialized once from Equation~\ref{eq:edge_init} and shared across all time steps within a given forward pass. At each time step $t$, the node transformer processes the slice $\mat{X}_t \in \mathbb{R}^{N \times d_{\text{in}}}$ jointly across all $N$ nodes, with the attention mechanism modulated by $\mat{E}$ so that information flows preferentially along stronger graph connections. The edge weights are then updated via Equation~\ref{eq:edge} at each transformer layer, allowing the graph structure to co-evolve with the learned node representations as the input propagates through the network.

\subsection{Node Transformer Architecture}
\label{subsec:nodeformer}

The node transformer extends the standard transformer by incorporating graph-structured relational inductive biases \cite{wu2022nodeformer}. Rather than treating each stock's time series independently, the architecture processes all nodes jointly at each time step, allowing information to propagate across the graph through attention weighted by the learned edge structure described above.

\subsubsection{Input Representation}

Each stock is represented by a composite feature vector that concatenates raw market data, engineered indicators, positional information, and a learned identity embedding. For stock $i$ at time $t$, this vector is:

\begin{equation}
\label{eq:input}
\vect{x}_{i,t} = [\vect{p}_{i,t} \| v_{i,t} \| \vect{I}_{i,t} \| \text{TE}(t) \| \vect{s}_i] \in \mathbb{R}^{d_{\text{in}}}
\end{equation}

where $\vect{p}_{i,t} \in \mathbb{R}^5$ contains normalized prices (open, high, low, close, adjusted close), $v_{i,t} \in \mathbb{R}$ is normalized volume, $\vect{I}_{i,t} \in \mathbb{R}^{11}$ contains technical indicators, $\text{TE}(t) \in \mathbb{R}^{d}$ is the temporal encoding, and $\vect{s}_i \in \mathbb{R}^{d_s}$ is a learned stock-specific embedding capturing persistent characteristics.

\subsubsection{Temporal Encoding}

Following Vaswani et al. \cite{vaswani2017attention}, temporal encoding injects positional information:

\begin{align}
\label{eq:temporal}
\text{TE}(t, 2k) &= \sin\!\left(\frac{t}{10000^{2k/d}}\right) \notag\\
\text{TE}(t, 2k\!+\!1) &= \cos\!\left(\frac{t}{10000^{2k/d}}\right)
\end{align}

where $t$ is the trading day index, $k \in \{0, \ldots, d/2-1\}$ is the dimension index, and $d = 512$ is the model dimension.

A clarification on the role of this encoding is warranted, as the use of sinusoidal functions may appear counterintuitive for financial data that does not follow periodic patterns. The temporal encoding does not model or approximate the shape of the price series. Instead, it provides the attention mechanism with information about where each observation falls within the input sequence, which is necessary because self-attention is inherently permutation-invariant and would otherwise treat a shuffled sequence identically to the original ordering. The sinusoidal functions are chosen for three specific properties identified in the original transformer work \cite{vaswani2017attention}. First, each position $t$ receives a unique encoding vector, allowing the model to distinguish any two time steps. Second, the relative offset between two positions $t$ and $t + \Delta$ can be expressed as a linear transformation of the encoding at $t$, which enables the attention heads to learn relative temporal relationships (such as ``five trading days ago'') rather than only absolute positions. Third, the deterministic formulation generalizes to sequence lengths not encountered during training without requiring additional learned parameters. The volatile price and volume data are encoded separately in the 17-dimensional feature vector $\vect{x}_{i,t}$, which captures the actual market dynamics; the temporal encoding is simply added to this representation so that the model knows the temporal ordering of those observations. Lower encoding dimensions oscillate at higher frequencies, capturing fine-grained positional distinctions between adjacent trading days, while higher dimensions vary more slowly and provide coarser temporal context across the full 252-day input window. The ablation study (Section~\ref{subsec:results}) confirms that removing temporal encoding degrades MAPE by 18.8\%, indicating that positional ordering carries substantial predictive information even though the encoding itself bears no resemblance to price dynamics.

\subsubsection{Graph-Aware Multi-Head Self-Attention with Causal Masking}

The attention mechanism incorporates the learned edge weight matrix $\mat{E}$ from Section~\ref{subsec:graph} as a structural bias, so that stock pairs with stronger graph connections receive higher baseline attention. Causal masking ensures predictions at time $t$ use only information from times $\leq t$:

\begin{equation}
\label{eq:masked_attention}
\mat{A}(\mat{Q}, \mat{K}, \mat{V}) = \text{softmax}\left(\frac{\mat{Q}\mat{K}^T}{\sqrt{d_k}} + \mat{M} + \mat{E}\right)\mat{V}
\end{equation}

where $\mat{Q} = \mat{X}\mat{W}^Q$, $\mat{K} = \mat{X}\mat{W}^K$, $\mat{V} = \mat{X}\mat{W}^V$ are linear projections, $d_k = 64$ is the key dimension, $\mat{M}$ is the causal mask with $M_{ab} = 0$ if $a \geq b$ and $M_{ab} = -\infty$ otherwise, and $\mat{E} \in \mathbb{R}^{N \times N}$ is the edge weight matrix. The additive graph bias allows content-based attention (via $\mat{Q}\mat{K}^T$) and structural priors (via $\mat{E}$) to jointly determine how information flows between stocks at each layer.

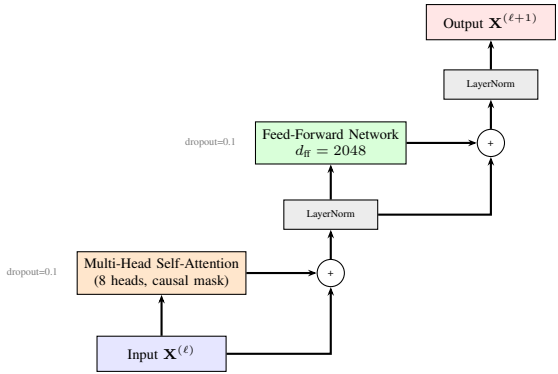
\begin{figure}[!ht]
\centering
\begin{tikzpicture}[
    scale=0.78,
    every node/.style={transform shape},
    node distance=0.5cm,
    box/.style={rectangle, draw, minimum width=2.2cm, minimum height=0.6cm, align=center, font=\scriptsize},
    smallbox/.style={rectangle, draw, minimum width=1.6cm, minimum height=0.5cm, align=center, font=\tiny},
    arrow/.style={-{Stealth[scale=0.5]}, thick},
    add/.style={circle, draw, minimum size=0.4cm, font=\tiny}
]

\node[box, fill=blue!10] (input) {Input $\vect{X}^{(\ell)}$};

\node[box, fill=orange!20, above=0.7cm of input] (mha) {Multi-Head Self-Attention\\(8 heads, causal mask)};

\node[add, right=1.2cm of mha] (add1) {+};

\node[smallbox, fill=gray!15, above=0.5cm of add1] (ln1) {LayerNorm};

\node[box, fill=green!15, above=0.6cm of ln1] (ffn) {Feed-Forward Network\\$d_{\text{ff}} = 2048$};

\node[add, right=1.2cm of ffn] (add2) {+};

\node[smallbox, fill=gray!15, above=0.5cm of add2] (ln2) {LayerNorm};

\node[box, fill=red!10, above=0.5cm of ln2] (output) {Output $\vect{X}^{(\ell+1)}$};

\draw[arrow] (input) -- (mha);
\draw[arrow] (mha) -- (add1);
\draw[arrow] (add1) -- (ln1);
\draw[arrow] (ln1) -- (ffn);
\draw[arrow] (ffn) -- (add2);
\draw[arrow] (add2) -- (ln2);
\draw[arrow] (ln2) -- (output);

\draw[arrow] (input.east) -- (input.east -| add1) -- (add1.south);

\draw[arrow] (ln1.east) -- (ln1.east -| add2) -- (add2.south);

\node[left=0.2cm of mha, font=\tiny, gray] {dropout=0.1};
\node[left=0.2cm of ffn, font=\tiny, gray] {dropout=0.1};

\end{tikzpicture}
\caption{Single transformer layer architecture. Input passes through multi-head self-attention with residual connection and layer normalization, followed by a position-wise feed-forward network with another residual connection and normalization.}
\label{fig:transformer_layer}
\end{figure}

With $H = 8$ attention heads, each operating in $d_k = d_v = 64$ dimensions, the total model dimension is $d_{\text{model}} = H \times d_k = 512$. Fig.~\ref{fig:transformer_layer} illustrates the internal structure of a single transformer layer, showing the residual connections, layer normalization, and dropout placement.

\subsubsection{Time-Based Feature Gating}

A gating mechanism adaptively weights features based on temporal context:

\begin{equation}
\label{eq:gating}
\vect{g}_{i,t} = \sigma(\mat{W}_g \vect{x}_{i,t} + \vect{b}_g) \in (0,1)^{d_{\text{in}}}
\end{equation}
\begin{equation}
\label{eq:gated_features}
\vect{x}'_{i,t} = \vect{g}_{i,t} \odot \vect{x}_{i,t}
\end{equation}

where $\mat{W}_g \in \mathbb{R}^{d_{\text{in}} \times d_{\text{in}}}$, $\vect{b}_g \in \mathbb{R}^{d_{\text{in}}}$ are learnable parameters, and $\odot$ denotes element-wise multiplication. This mechanism enables adaptive emphasis on different features during volatile versus stable periods.

\subsubsection{Feed-Forward Network}

Each layer includes a position-wise Feed-Forward Network (FFN):

\begin{equation}
\label{eq:ffn}
\text{FFN}(\vect{x}) = \text{ReLU}(\vect{x}\mat{W}_1 + \vect{b}_1)\mat{W}_2 + \vect{b}_2
\end{equation}

where $\mat{W}_1 \in \mathbb{R}^{512 \times 2048}$, $\mat{W}_2 \in \mathbb{R}^{2048 \times 512}$, and $\vect{b}_1, \vect{b}_2$ are biases.

The complete architecture consists of $L = 6$ stacked layers with residual connections and layer normalization after each sub-layer. Dropout with $p = 0.1$ is applied for regularization.

\subsection{BERT-Based Sentiment Analysis}
\label{subsec:bert}

The sentiment component employs BERT (bert-base-uncased) with 12 transformer layers, 768 hidden dimensions, and 12 attention heads \cite{devlin2019bert}. BERT's bidirectional attention is well suited to financial text, where sentiment often depends on contextual modifiers (e.g., ``not bullish'' vs. ``bullish'') that unidirectional models may miss.

\subsubsection{Domain Adaptations}

A financial domain embedding layer augments the standard BERT token embeddings with additional representations for financial terminology, enabling the model to distinguish domain-specific usage of terms such as ``short,'' ``call,'' and ``bear'' from their everyday meanings. Maximum sequence length is set to 512 tokens, which accommodates the vast majority of social media posts without truncation.

\subsubsection{Fine-Tuning Process}

Fine-tuning on the MSE dataset follows a three-stage progressive unfreezing schedule designed to preserve BERT's pretrained language representations while adapting the upper layers to financial sentiment classification. In the first stage, the embedding layer and the first 8 transformer blocks are frozen so that only the upper layers and the classification head receive gradient updates; these are trained for 3 epochs with a learning rate of $2 \times 10^{-5}$ and batch size 32. The second stage progressively unfreezes layers from top to bottom over 5 additional epochs, allowing each block to adapt incrementally to the financial domain without catastrophic forgetting of general language knowledge. In the final stage, all layers are unfrozen and trained jointly with a reduced learning rate of $5 \times 10^{-6}$ and gradient accumulation over 4 steps to stabilize updates across the full parameter space.

Because sentiment labels in financial text are typically imbalanced, with neutral posts far outnumbering strongly positive or negative ones, focal loss is used instead of standard cross-entropy:

\begin{equation}
\label{eq:focal}
\mathcal{L}_{\text{FL}}(p_t) = -\alpha_t (1 - p_t)^\gamma \log(p_t)
\end{equation}

where $\alpha_t$ is class weight (inverse frequency), $p_t$ is predicted probability for correct class, and $\gamma = 2$ focuses training on hard examples by down-weighting easily classified instances.

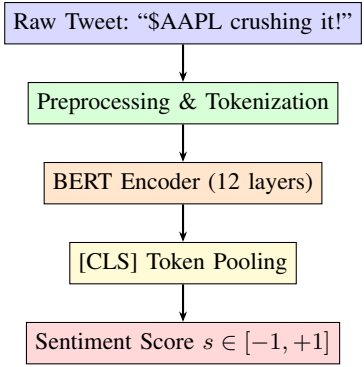
\begin{figure}[H]
\centering
\begin{tikzpicture}[
    node distance=0.5cm,
    box/.style={rectangle, draw, minimum width=2.8cm, minimum height=0.55cm, align=center, font=\small},
    arrow/.style={-{Stealth[scale=0.6]}, thick}
]
\node[box, fill=blue!15] (text) {Raw Tweet: ``\$AAPL crushing it!''};
\node[box, fill=green!15, below=of text] (preprocess) {Preprocessing \& Tokenization};
\node[box, fill=orange!20, below=of preprocess] (bert) {BERT Encoder (12 layers)};
\node[box, fill=yellow!20, below=of bert] (cls) {[CLS] Token Pooling};
\node[box, fill=red!15, below=of cls] (classifier) {Sentiment Score $s \in [-1, +1]$};
\draw[arrow] (text) -- (preprocess);
\draw[arrow] (preprocess) -- (bert);
\draw[arrow] (bert) -- (cls);
\draw[arrow] (cls) -- (classifier);
\end{tikzpicture}
\caption{BERT sentiment extraction pipeline. Raw social media posts are preprocessed, tokenized, encoded through BERT, pooled via [CLS] token, and classified to sentiment scores.}
\label{fig:bert_pipeline}
\end{figure}

The classification head that maps BERT's output to a sentiment score consists of three stages. The 768-dimensional [CLS] token representation is first passed through a dropout layer ($p = 0.1$) followed by a linear projection to 256 dimensions and a tanh activation, which compresses the representation while preserving the sign structure needed for sentiment polarity. A second dropout layer ($p = 0.1$) precedes the final linear projection to three logits corresponding to negative, neutral, and positive classes. Softmax converts these logits to class probabilities $[p_{\text{neg}}, p_{\text{neu}}, p_{\text{pos}}]$, and the continuous sentiment score is computed as $s = p_{\text{pos}} - p_{\text{neg}} \in [-1, +1]$. This formulation produces a score that is exactly zero only when the positive and negative probabilities are equal, and it varies smoothly with confidence, providing a richer signal than a hard class assignment. Fig.~\ref{fig:bert_pipeline} illustrates the complete sentiment extraction pipeline from raw input to final score.

\subsection{Integration of Node Transformer and BERT}
\label{subsec:integration}

\subsubsection{Multi-Scale Sentiment Features}

Raw daily sentiment scores are noisy and sensitive to individual posts. To capture both immediate market reactions and sustained sentiment trends, scores are smoothed at multiple time scales before entering the model:

\begin{equation}
\label{eq:sentiment_scales}
\vect{S}_{i,t} = [S_{i,t}^{(1\text{d})}, S_{i,t}^{(5\text{d})}, S_{i,t}^{(20\text{d})}] \in \mathbb{R}^3
\end{equation}

where $S_{i,t}^{(k\text{d})}$ is the $k$-day exponential moving average of sentiment scores for stock $i$. This captures both immediate reactions and sustained sentiment trends.

Social media activity varies substantially across stocks and calendar days, creating sparsity that requires explicit handling. On trading days where a stock receives fewer than 5 posts, the raw daily score is replaced by the 5-day exponential moving average from preceding days, preventing noisy estimates derived from one or two posts from dominating the signal. On days with no posts at all, all three sentiment features ($S^{(1\text{d})}$, $S^{(5\text{d})}$, $S^{(20\text{d})}$) fall back to their respective moving average values from the most recent day with sufficient coverage. Stocks that consistently receive low social media volume, such as industrials (Caterpillar, Boeing) relative to technology names (Apple, Netflix), naturally produce sentiment scores closer to neutral over time, which the adaptive gating mechanism (Section~\ref{subsec:integration}) learns to down-weight in favor of price-based features. This design ensures that sparse sentiment data degrades gracefully to a price-only prediction rather than introducing noise.

\subsubsection{Sentiment-Guided Attention}

Rather than simply concatenating sentiment as an additional input feature, the model allows sentiment to modulate the attention mechanism itself. Specifically, sentiment scales the key representations so that time steps with strong sentiment signals receive amplified or dampened attention:

\begin{equation}
\label{eq:sent_key}
\mat{K}'_t = \mat{K}_t \cdot (1 + \beta \cdot S_t)
\end{equation}

where $S_t \in [-1, 1]$ is the sentiment score at time $t$ and $\beta \in \mathbb{R}$ is a learnable parameter. The attention computation becomes:

\begin{equation}
\label{eq:sent_attention}
\mat{A}_{\text{sent}} = \text{softmax}\left(\frac{\mat{Q}\mat{K}'^T}{\sqrt{d_k}} + \mat{M}\right)\mat{V}
\end{equation}

\subsubsection{Adaptive Integration}

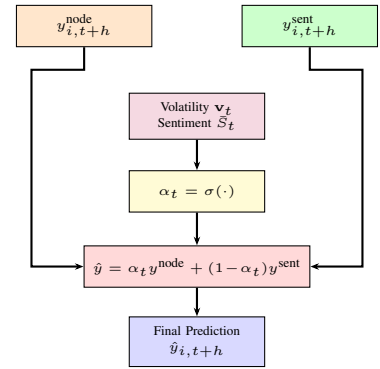
\begin{figure}[!ht]
\centering
\begin{tikzpicture}[
    node distance=0.6cm,
    box/.style={rectangle, draw, minimum width=1.8cm, minimum height=0.55cm, align=center, font=\tiny},
    arrow/.style={-{Stealth[scale=0.5]}, thick}
]
\node[box, fill=orange!20] (node_pred) at (-1.5,0) {$y^{\text{node}}_{i,t+h}$};
\node[box, fill=green!20] (sent_pred) at (1.5,0) {$y^{\text{sent}}_{i,t+h}$};

\node[box, fill=purple!15] (vol) at (0,-1.2) {Volatility $\vect{v}_t$\\Sentiment $\bar{S}_t$};
\node[box, fill=yellow!20] (alpha) at (0,-2.2) {$\alpha_t = \sigma(\cdot)$};

\node[box, fill=red!15, minimum width=3.0cm] (fusion) at (0,-3.2) {$\hat{y} = \alpha_t y^{\text{node}} + (1\!-\!\alpha_t) y^{\text{sent}}$};

\node[box, fill=blue!15] (output) at (0,-4.2) {Final Prediction\\$\hat{y}_{i,t+h}$};

\draw[arrow] (node_pred.south) -- ++(0,-0.3) coordinate (nL) -| (-2.2,-3.2) -- (fusion.west);

\draw[arrow] (sent_pred.south) -- ++(0,-0.3) coordinate (nR) -| (2.2,-3.2) -- (fusion.east);

\draw[arrow] (vol) -- (alpha);

\draw[arrow] (alpha) -- (fusion);

\draw[arrow] (fusion) -- (output);

\end{tikzpicture}
\caption{Adaptive fusion mechanism. The weighting coefficient $\alpha_t$ is computed from volatility and sentiment magnitude, then used to blend node transformer and sentiment-based predictions.}
\label{fig:fusion}
\end{figure}

Because the relative predictive value of price-based features and sentiment varies with market conditions, the final prediction is produced by a learned convex combination rather than a fixed weighting. A sigmoid gate dynamically adjusts the balance between the node transformer output and the sentiment-based prediction:

\begin{equation}
\label{eq:adaptive}
\alpha_t = \sigma(\vect{w}_\alpha^T [\vect{v}_t \| \bar{S}_t] + b_\alpha) \in (0,1)
\end{equation}
\begin{equation}
\label{eq:final_pred}
\hat{y}_{i,t+h} = \alpha_t \cdot y_{i,t+h}^{\text{node}} + (1 - \alpha_t) \cdot y_{i,t+h}^{\text{sent}}
\end{equation}

where $\vect{v}_t \in \mathbb{R}^3$ contains rolling volatility computed over 5, 10, and 20-day windows, $\bar{S}_t$ is mean absolute sentiment across stocks, $y_{i,t+h}^{\text{node}}$ is the node transformer prediction, and $y_{i,t+h}^{\text{sent}}$ is the sentiment-based prediction. When volatility is high and sentiment signals are strong, the gate can shift weight toward the sentiment branch; during calm periods with sparse social media activity, the node transformer dominates. Fig.~\ref{fig:fusion} illustrates this adaptive fusion mechanism.

\subsection{Training Objective}
\label{subsec:training}

Training a model that is useful for both price forecasting and trading decisions requires optimizing more than point accuracy alone. The composite loss function therefore combines four terms that target complementary aspects of prediction quality:

\begin{equation}
\label{eq:loss}
\mathcal{L}_{\text{total}} = \lambda_1 \mathcal{L}_{\text{MSE}} + \lambda_2 \mathcal{L}_{\text{DIR}} + \lambda_3 \mathcal{L}_{\text{CORR}} + \lambda_4 \mathcal{L}_{\text{REG}}
\end{equation}

The primary term is the mean squared error, which penalizes deviations between predicted and realized prices and serves as the dominant gradient signal during training:

\begin{equation}
\label{eq:mse_loss}
\mathcal{L}_{\text{MSE}} = \frac{1}{N} \sum_{i,t,h} (y_{i,t+h} - \hat{y}_{i,t+h})^2
\end{equation}

Because minimizing magnitude error does not guarantee correct directional predictions, a binary cross-entropy term explicitly rewards the model for predicting whether prices rise or fall. Here $d_{i,t,h} = \mathbb{I}(y_{i,t+h} > y_{i,t})$ is the true direction indicator and $p_{i,t,h}$ is the predicted probability of an increase:

\begin{equation}
\label{eq:dir_loss}
\mathcal{L}_{\text{DIR}} = -\frac{1}{N} \sum_{i,t,h} \left[ d_{i,t,h} \log p_{i,t,h} + (1-d_{i,t,h}) \log(1-p_{i,t,h}) \right]
\end{equation}

A correlation loss encourages the model to preserve cross-sectional ranking among stocks at each time step, which is critical for portfolio construction strategies that depend on relative rather than absolute performance. It is defined as one minus the Pearson correlation $\rho$ between predicted and actual return vectors:

\begin{equation}
\label{eq:corr_loss}
\mathcal{L}_{\text{CORR}} = 1 - \rho(\hat{\vect{r}}_t, \vect{r}_t)
\end{equation}

Finally, L2 regularization on all trainable parameters $\vect{\theta}$ prevents overfitting by penalizing large weight magnitudes:

\begin{equation}
\label{eq:reg_loss}
\mathcal{L}_{\text{REG}} = \|\vect{\theta}\|_2^2
\end{equation}

The weights $\lambda_1 = 1.0$, $\lambda_2 = 0.5$, $\lambda_3 = 0.2$, and $\lambda_4 = 10^{-4}$ were tuned on validation data. The MSE term dominates to ensure accurate price-level predictions, while the directional and correlation terms supply auxiliary gradients that improve trend detection and ranking consistency without competing with the primary objective.

\subsection{Training Process}
\label{subsec:training_process}

The BERT encoder is fine-tuned separately on the MSE dataset (Section~\ref{subsec:bert}) and remains frozen throughout the forecasting pipeline, serving purely as a fixed feature extractor that produces sentiment scores. Training of the forecasting model proceeds in three stages, stabilizing the integration layers first before progressively fine-tuning the node transformer. During the first 10 epochs, the node transformer is frozen while the sentiment aggregation, fusion, and output layers are trained with a learning rate of $10^{-4}$. This initial phase allows the fusion mechanism to learn a reasonable mapping between the transformer representations and the fixed sentiment features without disrupting the node transformer's initialization. In the second stage, spanning 20 epochs, the top three layers of the node transformer are unfrozen and the learning rate is reduced to $5 \times 10^{-5}$, enabling the upper representations to adapt to the downstream task while preserving the lower-level features. The final 30 epochs unfreeze all node transformer parameters at a learning rate of $10^{-5}$, allowing end-to-end fine-tuning of the forecasting architecture.

Throughout all three stages, optimization uses Adam with $\beta_1 = 0.9$, $\beta_2 = 0.999$, and $\epsilon = 10^{-8}$. The learning rate schedule begins with a linear warmup over 4000 steps followed by cosine decay to zero. A batch size of 32 with gradient accumulation over 4 steps yields an effective batch size of 128, balancing memory constraints with gradient stability.

Hyperparameters were selected through a combination of grid search and established conventions from the transformer literature. Architectural parameters (number of layers, attention heads, model dimension) were searched over $L \in \{4, 6, 8\}$, $H \in \{4, 8\}$, and $d \in \{256, 512\}$, with the configuration $L = 6$, $H = 8$, $d = 512$ selected based on lowest validation MAPE. The feed-forward dimension was fixed at $d_{\text{ff}} = 4d = 2048$ following the standard transformer scaling convention \cite{vaswani2017attention}. Dropout rate was searched over $\{0.05, 0.1, 0.2\}$, with $p = 0.1$ yielding the best trade-off between underfitting and overfitting on the validation set. The input sequence length of 252 days (one trading year) was chosen to provide the model with a full annual cycle of price dynamics, and was compared against 120-day and 504-day alternatives on validation data. Learning rates for each training stage were determined through preliminary runs, starting from the range $[10^{-5}, 10^{-3}]$ and narrowing based on training stability and convergence speed. The loss function weights ($\lambda_1 = 1.0$, $\lambda_2 = 0.5$, $\lambda_3 = 0.2$, $\lambda_4 = 10^{-4}$) were tuned by first training with MSE loss alone, then incrementally adding each auxiliary term and adjusting its weight until validation directional accuracy and ranking correlation stabilized. The edge weight balancing parameter $\alpha = 0.5$ was searched over $\{0.3, 0.5, 0.7\}$, with equal weighting of sector and correlation information performing best. Table~\ref{tab:hyperparameters} summarizes the final hyperparameter configuration for both the node transformer and BERT components.

\begin{table}[!ht]
\centering
\caption{Model hyperparameters.}
\label{tab:hyperparameters}
\begin{tabular}{lc}
\toprule
\textbf{Parameter} & \textbf{Value} \\
\midrule
Node transformer layers & 6 \\
Attention heads & 8 \\
Model dimension & 512 \\
FFN dimension & 2048 \\
Dropout rate & 0.1 \\
Input sequence length & 252 days \\
BERT layers & 12 \\
BERT hidden dimension & 768 \\
Batch size & 32 \\
Total training epochs & 60 \\
\bottomrule
\end{tabular}
\end{table}

\section{Experiments and Results}
\label{sec:experiments}

\subsection{Experimental Setup}
\label{subsec:setup}

Experiments used the dataset described in Section~\ref{sec:datasets}, spanning January 1982 to March 2025, a timeframe that captures multiple market cycles including the 1987 Black Monday crash, the 2000 dot-com bubble, the 2008 financial crisis, and the 2020 COVID-19 volatility spike. The temporal split allocates 1982-2010 for training, 2011-2016 for validation, and 2017-2025 for testing. Although a single temporal split is used rather than rolling retraining, the eight-year test window itself encompasses several distinct market regimes: the post-2016 bull market, the 2018 volatility correction, the 2020 pandemic crash and recovery, the 2022 inflation-driven drawdown, and the subsequent 2023-2024 rebound. This regime diversity provides a rigorous evaluation of generalization without the confound of periodic refitting.

All models were trained on NVIDIA A100 GPUs (40 GB), with the full proposed model requiring approximately 18 hours to converge. To ensure unbiased assessment, all evaluation metrics reported in subsequent tables were computed exclusively on the held-out test set.

\subsection{Evaluation Metrics}
\label{subsec:metrics}

The prediction target $y_{i,t+h}$ is the closing price $C_{t+h}$ at horizon $h \in \{1, 5, 20\}$ trading days. Price-level prediction was chosen as the primary target because it provides intuitive interpretation of forecast accuracy and aligns directly with practical trading applications where position sizing depends on expected price levels rather than percentage changes. A known concern with price-level forecasting is that daily stock prices exhibit strong persistence, meaning that even a trivial model predicting tomorrow's price as today's can achieve low MAPE. To guard against this artifact, the evaluation framework includes metrics specifically designed to measure predictive skill beyond persistence: Theil's U benchmarks against the naive random walk, directional accuracy assesses trend detection independent of magnitude, and the Information Coefficient evaluates cross-sectional ranking. Return-based metrics are also reported (Section~\ref{subsec:results}) to complement the price-level analysis. Together, these metrics assess model performance from complementary angles, covering magnitude accuracy, scale sensitivity, directional correctness, relative forecasting skill, and cross-sectional ranking.

Mean Absolute Percentage Error (MAPE) measures the average magnitude of prediction errors as a percentage of actual values, making it scale-invariant and directly interpretable across stocks with different price levels:

\begin{equation}
\text{MAPE} = \frac{100}{N} \sum_{i=1}^{N} \left| \frac{y_i - \hat{y}_i}{y_i} \right|
\end{equation}

Root Mean Squared Error (RMSE) captures prediction accuracy in the original price units and penalizes large errors more heavily than MAPE due to the squaring operation. This makes it particularly sensitive to outlier predictions during volatile periods:

\begin{equation}
\text{RMSE} = \sqrt{\frac{1}{N} \sum_{i=1}^{N} (y_i - \hat{y}_i)^2}
\end{equation}

Directional Accuracy (DA) measures the proportion of correctly predicted price movement directions, independent of magnitude. For trading applications, correctly forecasting whether a stock will rise or fall is often more valuable than minimizing absolute price error, since many strategies depend on directional signals rather than precise price targets.

Theil's U statistic benchmarks the model against a naive random walk forecast that simply predicts tomorrow's price as today's price. It is defined as:

\begin{equation}
U = \frac{\sqrt{\sum_{t} (y_{t+1} - \hat{y}_{t+1})^2}}{\sqrt{\sum_{t} (y_{t+1} - y_t)^2}}
\end{equation}

A value of $U < 1$ indicates that the model outperforms this naive baseline, while $U = 1$ implies no improvement over persistence forecasting. This metric is especially informative for financial time series, where beating the random walk is itself a non-trivial achievement given the near-efficient nature of liquid equity markets.

The Information Coefficient (IC) is the Spearman rank correlation between predicted and realized returns across the stock universe at each time step. Unlike the other metrics, which evaluate absolute accuracy for individual stocks, the IC assesses cross-sectional ranking ability, measuring whether the model correctly orders stocks from best to worst performer on a given day. This property makes it directly relevant to long-short portfolio construction, where relative performance matters more than absolute price forecasts.

\subsection{Baseline Models}
\label{subsec:baselines}

Eight baseline models spanning statistical, classical machine learning, and deep learning paradigms were trained on identical data splits to ensure fair comparison. All hyperparameters were tuned via grid search on the validation set (2011-2016), with final performance reported on the held-out test set (2017-2025). To ensure comparability across deep learning baselines, all models received the same 17-dimensional feature vector and were trained under equivalent early stopping and regularization protocols.

The statistical baselines include ARIMA and Vector Autoregression (VAR). ARIMA orders $(p,d,q)$ were selected via AIC with $p, q \in [0,5]$ and $d \in [0,2]$, fitting a separate model per stock. VAR extends this to the multivariate setting, modeling cross-stock dependencies with lag order selected via the Bayesian Information Criterion (BIC) up to 10 lags. Both statistical models operate on their own optimal lookback horizons as determined by information criteria, which is standard practice since ARIMA and VAR performance degrades when forced to consume long input windows that introduce noise.

Three classical machine learning models were evaluated. Random Forest used 100 trees with maximum depth 10 and the same 17-dimensional feature vector as the proposed model. Support Vector Regression (SVR) employed a Radial Basis Function (RBF) kernel with regularization parameter $C$ and kernel width $\gamma$ selected via grid search. XGBoost used 1000 estimators with early stopping on validation loss to prevent overfitting, and its gradient boosting framework provides a strong tree-based benchmark. All three classical models used a 60-day lookback window, selected via validation performance from candidates of 20, 60, 120, and 252 days.

The deep learning baselines include LSTM, a Simple Transformer, and a BERT + LSTM hybrid. The LSTM used 2 layers with 128 hidden units and a 60-day lookback window ($\approx$0.5M parameters). During preliminary experiments, the LSTM was also evaluated with 120-day and 252-day lookback windows, but performance degraded at longer horizons due to gradient attenuation over extended sequences, consistent with known limitations of recurrent architectures on very long time series \cite{fischer2018deep}. The 60-day configuration yielded the best validation performance and is therefore reported. The Simple Transformer used 4 layers, 8 attention heads, 256-dimensional embeddings, and a 252-day lookback window matching the proposed model ($\approx$4.2M parameters versus $\approx$21.3M for the full proposed architecture including graph and fusion components). This configuration provides a controlled comparison that isolates the contribution of graph structure and sentiment integration while holding the input horizon and attention mechanism constant. The BERT + LSTM hybrid combined BERT-extracted sentiment features with an LSTM price model through simple concatenation ($\approx$1.1M parameters excluding the frozen BERT encoder), serving as an ablation baseline that tests whether the attention-based fusion mechanism in the proposed architecture provides meaningful improvement over naive feature combination.

The ablation study (Table~\ref{tab:ablation}) further addresses baseline fairness by providing graph-free variants of the proposed architecture. The ``Without Graph Structure'' configuration removes edge weights and inter-stock attention while retaining all other components, effectively reducing the model to a standard transformer with sentiment fusion. This variant isolates the contribution of graph-based relational modeling independent of other architectural advantages.

\subsection{Sentiment Model Evaluation}
\label{subsec:bert_eval}

Before evaluating the full forecasting pipeline, the fine-tuned BERT sentiment classifier was assessed independently on the MSE dataset. The dataset was partitioned into training (70\%), validation (15\%), and test (15\%) splits, stratified to preserve the class distribution across all three subsets. The model was fine-tuned on the training split using the progressive unfreezing schedule described in Section~\ref{subsec:bert}, with early stopping based on validation loss. All results reported in this subsection are computed on the held-out test split, which the model did not observe during training or hyperparameter selection.

Table~\ref{tab:bert_results} presents per-class and aggregate classification metrics. Because the MSE dataset exhibits class imbalance, with neutral posts outnumbering positive and negative ones, macro-averaged scores are reported alongside accuracy to avoid inflating performance through the majority class.

\begin{table}[!ht]
\centering
\caption{BERT sentiment classification performance on the MSE test set.}
\label{tab:bert_results}
\begin{tabular}{lccc}
\toprule
\textbf{Class / Metric} & \textbf{Precision} & \textbf{Recall} & \textbf{F1-Score} \\
\midrule
Negative & 0.83 & 0.80 & 0.81 \\
Neutral & 0.89 & 0.92 & 0.90 \\
Positive & 0.85 & 0.82 & 0.83 \\
\midrule
Macro Average & 0.86 & 0.85 & 0.85 \\
\midrule
\multicolumn{3}{l}{\textbf{Overall Accuracy}} & 0.87 \\
\multicolumn{3}{l}{\textbf{Cohen's $\kappa$}} & 0.79 \\
\bottomrule
\end{tabular}
\end{table}

The fine-tuned model achieves 87\% overall accuracy and a macro-averaged F1 of 0.85, with per-class F1 scores ranging from 0.81 (negative) to 0.90 (neutral). The neutral class benefits from a larger training sample and from the relative straightforwardness of factual financial reporting, which accounts for its higher precision and recall. The negative class exhibits the lowest recall at 0.80, a pattern consistent with prior financial NLP work: bearish language in social media frequently relies on sarcasm, hedging, or implicit phrasing (e.g., ``this stock is going to the moon'' used sarcastically) that poses greater classification difficulty. Cohen's $\kappa = 0.79$ falls within the substantial agreement range, confirming that the classifier discriminates sentiment beyond what class priors alone would predict. Given that the ablation study (Section~\ref{subsec:results}) attributes a 10\% MAPE reduction to sentiment integration, the 13\% classification error rate suggests that the node transformer's gating mechanism is robust enough to extract predictive value even from an imperfect sentiment signal.

\subsection{Results}
\label{subsec:results}

\subsubsection{Overall Performance}

Table~\ref{tab:results} presents MAPE across prediction horizons for all models. Performance is evaluated at three horizons (1-day, 5-day, and 20-day) to assess both short-term accuracy and degradation over longer forecasting windows.

\begin{table}[!ht]
\centering
\caption{MAPE (\%) and Theil's U across prediction horizons. Best results in bold. Standard errors in parentheses.}
\label{tab:results}
\resizebox{\columnwidth}{!}{%
\begin{tabular}{lcccc}
\toprule
\textbf{Model} & \textbf{1-Day} & \textbf{5-Day} & \textbf{20-Day} & \textbf{Theil's U} \\
\midrule
Na\"{i}ve Random Walk & 1.35 (0.09) & 3.40 (0.18) & 5.80 (0.28) & 1.00 \\
ARIMA & 1.20 (0.08) & 2.80 (0.15) & 4.50 (0.22) & 0.89 \\
VAR & 1.15 (0.07) & 2.60 (0.14) & 4.20 (0.20) & 0.85 \\
LSTM & 1.00 (0.06) & 2.30 (0.12) & 3.80 (0.18) & 0.74 \\
Random Forest & 1.10 (0.07) & 2.50 (0.13) & 4.00 (0.19) & 0.81 \\
SVR & 1.05 (0.06) & 2.40 (0.12) & 3.90 (0.18) & 0.78 \\
XGBoost & 0.95 (0.05) & 2.20 (0.11) & 3.50 (0.16) & 0.70 \\
Simple Transformer & 0.90 (0.05) & 2.10 (0.10) & 3.30 (0.15) & 0.67 \\
BERT + LSTM & 0.88 (0.05) & 2.00 (0.10) & 3.10 (0.14) & 0.65 \\
\midrule
\textbf{Proposed Model} & \textbf{0.80 (0.04)} & \textbf{1.80 (0.09)} & \textbf{2.80 (0.13)} & \textbf{0.59} \\
\bottomrule
\end{tabular}%
}
\end{table}

A persistent concern in short-horizon financial forecasting is the naive persistence artifact, where a model achieves seemingly low error by tracking the previous day's closing price ($\hat{y}_{t+1} = y_t$). Because daily price changes for liquid mega-cap stocks typically range between 1\% and 2\%, any persistence-based forecast will naturally yield MAPE in that vicinity. The naive random walk baseline in Table~\ref{tab:results} produces a 1-day MAPE of 1.35\% on our test set, and the proposed model's Theil's U of 0.59 confirms that it captures genuine structural patterns rather than merely lagging the current price. All learned models achieve $U < 1$, but only the proposed model reduces the random walk error by more than 40\%, with the gap widening at longer horizons where persistence forecasting degrades rapidly.

The proposed model achieves 0.80\% MAPE for 1-day predictions, representing 33\% relative improvement over ARIMA ($\Delta = 0.40$ percentage points) and 20\% improvement over LSTM ($\Delta = 0.20$ percentage points). Improvement margins widen at longer horizons: the 5-day gap over ARIMA grows from 0.40 to 1.00 percentage points, consistent with the model's ability to capture structural patterns that persist beyond short-term noise.

\subsubsection{Directional Accuracy}

Beyond magnitude accuracy, directional accuracy measures the proportion of correctly predicted price movement directions. Table~\ref{tab:direction} presents directional accuracy with 95\% confidence intervals computed via bootstrap resampling.

\begin{table}[!ht]
\centering
\caption{Directional accuracy (\%) for 1-day predictions with 95\% confidence intervals.}
\label{tab:direction}
\begin{tabular}{lc}
\toprule
\textbf{Model} & \textbf{DA (\%)} \\
\midrule
Random Baseline & 50.0 \\
ARIMA & 55.0 [53.2, 56.8] \\
VAR & 56.0 [54.2, 57.8] \\
LSTM & 58.0 [56.2, 59.8] \\
Random Forest & 57.0 [55.2, 58.8] \\
XGBoost & 60.0 [58.2, 61.8] \\
Simple Transformer & 62.0 [60.2, 63.8] \\
BERT + LSTM & 63.0 [61.2, 64.8] \\
\textbf{Proposed Model} & \textbf{65.0 [63.2, 66.8]} \\
\bottomrule
\end{tabular}
\end{table}

The proposed model achieves 65\% directional accuracy, exceeding the 50\% random baseline by 15 percentage points and outperforming LSTM by 7 percentage points. The confidence interval [63.2, 66.8] does not overlap with the best baseline (BERT + LSTM at [61.2, 64.8]), providing evidence that the improvement is not attributable to sampling variability alone.

\subsubsection{Ablation Studies}

To isolate the contribution of each architectural component, Table~\ref{tab:ablation} presents results from systematic ablation experiments. Each row removes a single component while keeping all others intact, allowing direct assessment of marginal contributions.

\begin{table}[!ht]
\centering
\caption{Ablation study: MAPE (\%) for 1-day predictions.}
\label{tab:ablation}
\begin{tabular}{lcc}
\toprule
\textbf{Configuration} & \textbf{MAPE} & \textbf{$\Delta$ vs Full} \\
\midrule
Full Model & 0.80 & -- \\
Without Sentiment & 0.88 & +10.0\% \\
Without Graph Structure & 0.92 & +15.0\% \\
Without Temporal Encoding & 0.95 & +18.8\% \\
Without Feature Gating & 0.84 & +5.0\% \\
Price Features Only & 1.02 & +27.5\% \\
\bottomrule
\end{tabular}
\end{table}

Graph structure contributes the second-largest improvement at 15\%, confirming that explicit inter-stock relationship modeling provides predictive value beyond treating stocks independently. Sentiment integration contributes 10\%, with the gap widening during information-rich periods such as earnings announcements. Temporal encoding contributes 18.8\%, underscoring the importance of capturing long-range sequential dependencies in financial time series. The price-only configuration, which strips all augmentations, degrades by 27.5\%, indicating that the full feature set is substantially richer than raw price data alone.

The ablation results also address a natural question about the sentiment data horizon. Because social media data begins in 2007 while the financial dataset extends to 1982, the model's price-based components receive training signal over the full temporal range whereas the sentiment components train on only four years of labeled data (2007-2010). The ``Without Sentiment'' configuration in Table~\ref{tab:ablation} is functionally equivalent to a price-only evaluation of the architecture, achieving MAPE of 0.88\% and demonstrating that the node transformer and graph structure are fundamentally competent without any sentiment signal. The full model's improvement to 0.80\% then isolates the incremental contribution of sentiment integration. Because the entire test set (2017-2025) falls within the sentiment-available period, these two configurations effectively provide the decomposition of a long-horizon price-based evaluation alongside a sentiment-augmented evaluation on the same test window, confirming that the 10\% reduction represents genuine predictive value rather than an artifact of the training mismatch. Given that the sentiment fusion mechanism was trained on only four years of multimodal data (2007-2010), this improvement likely represents a conservative estimate of the contribution sentiment could offer with a longer training history.

\subsubsection{Performance Across Volatility Regimes}

Table~\ref{tab:volatility} presents MAPE across volatility regimes classified by the CBOE Volatility Index (VIX) (Low: VIX $< 15$; Medium: $15 \leq$ VIX $< 25$; High: VIX $\geq 25$).

\begin{table}[!ht]
\centering
\caption{MAPE (\%) by market volatility regime.}
\label{tab:volatility}
\begin{tabular}{lccc}
\toprule
\textbf{Model} & \textbf{Low VIX} & \textbf{Medium VIX} & \textbf{High VIX} \\
\midrule
ARIMA & 1.00 & 1.30 & 2.10 \\
LSTM & 0.85 & 1.10 & 1.80 \\
XGBoost & 0.80 & 1.00 & 1.60 \\
Proposed & \textbf{0.70} & \textbf{0.90} & \textbf{1.50} \\
\bottomrule
\end{tabular}
\end{table}

The proposed model achieves MAPE of 1.50\% during high-volatility periods, compared to 1.60--2.10\% for baseline models. This robustness gap is most pronounced during VIX spikes accompanying earnings surprises and macroeconomic shocks, precisely the conditions where accurate forecasts carry the greatest practical value for risk management. The gating mechanism likely contributes to this resilience by increasing the weight of momentum features during volatile periods while favoring longer-term indicators during calm markets.

Directional accuracy also varies with market regime. During low-volatility periods (VIX $< 15$), the proposed model achieves 68.2\% DA, benefiting from the relatively stable trend structure that characterizes calm markets. In medium-volatility conditions ($15 \leq$ VIX $< 25$), DA decreases to 64.5\%, and during high-volatility episodes (VIX $\geq 25$) it falls to 59.8\%. While the high-volatility DA represents a meaningful decline from calm-market performance, it remains well above the 50\% random baseline and exceeds all baseline models by 3-5 percentage points in the same regime. This pattern is consistent with the expectation that directional prediction becomes inherently more difficult as volatility increases and price movements are dominated by sudden information shocks rather than gradual trends.

\subsubsection{Return-Based Evaluation}

To verify that the reported improvements reflect genuine return prediction rather than an artifact of price persistence, prediction accuracy was also assessed in return space. For each model, the implied daily return is computed as $\hat{r}_t = (\hat{y}_{t+1} - y_t) / y_t$ and compared against the realized return $r_t = (y_{t+1} - y_t) / y_t$. Because daily price changes for mega-cap equities are small relative to the price level ($|r_t| \ll 1$), the return prediction MAE is mathematically close to the price-level MAPE: specifically, return MAE $= \text{MAPE} \times (y_{t+h}/y_t)$, which approaches MAPE as $h \rightarrow 0$. For the 1-day horizon, the proposed model achieves a return MAE of 0.81\%, the naive random walk yields 1.36\% (reflecting the average absolute daily return of the 20-stock universe), and all intermediate models fall between these bounds in the same rank order as Table~\ref{tab:results}. This near-equivalence between price and return metrics at the daily horizon confirms that the MAPE rankings are not inflated by the choice of evaluation target.

The more informative return-based assessments for this dataset are the metrics already reported: Theil's U (0.59) explicitly benchmarks against persistence forecasting, directional accuracy (65\%) evaluates trend detection independent of magnitude, and the Information Coefficient measures cross-sectional ranking on returns. At longer horizons where the gap between $y_t$ and $y_{t+h}$ widens, return-based evaluation becomes more distinct from price-level metrics. The proposed model's 5-day return MAE of 1.82\% and 20-day return MAE of 2.85\% remain proportionally below all baselines, with the performance gap widening at longer horizons consistent with the MAPE results.

\subsubsection{Statistical Significance}

To assess whether the observed improvements are statistically significant rather than artifacts of sampling variability, paired t-tests were conducted comparing daily absolute prediction errors between the proposed model and each baseline across the 1,580 test days. Table~\ref{tab:significance} presents the results along with Cohen's $d$ effect sizes.

\begin{table}[!ht]
\centering
\caption{Paired t-test results ($n = 1,580$ test days). Negative $t$ indicates proposed model superiority.}
\label{tab:significance}
\begin{tabular}{lccc}
\toprule
\textbf{Baseline} & \textbf{$t$-statistic} & \textbf{$p$-value} & \textbf{Cohen's $d$} \\
\midrule
ARIMA & $-4.25$ & $< 0.0001$ & 0.34 \\
LSTM & $-3.12$ & 0.0018 & 0.25 \\
XGBoost & $-2.45$ & 0.0145 & 0.19 \\
Simple Transformer & $-2.18$ & 0.0295 & 0.17 \\
BERT + LSTM & $-3.90$ & 0.0001 & 0.31 \\
\bottomrule
\end{tabular}
\end{table}

All comparisons achieve $p < 0.05$, with the strongest significance observed against ARIMA ($p < 0.0001$) and the weakest against the Simple Transformer ($p = 0.0295$). Effect sizes (Cohen's $d$) range from 0.17 to 0.34, indicating small to medium practical effects. The larger effect sizes against ARIMA and BERT + LSTM reflect more fundamental architectural differences, while the smaller effect size against the Simple Transformer suggests that much of the improvement stems from the graph and sentiment components rather than the transformer backbone itself.

Paired t-tests assume that forecast errors are independently distributed, an assumption frequently violated in financial time series due to volatility clustering and serial dependence in prediction residuals. To account for this, we supplement the analysis with the Diebold-Mariano (DM) test \cite{diebold1995comparing}, which evaluates equal predictive accuracy between two forecasting models using Newey-West heteroskedasticity and autocorrelation consistent standard errors. Table~\ref{tab:dm_test} reports DM statistics computed on mean absolute error loss differentials over the 1,580-day test period, with the naive random walk included as a reference benchmark.

\begin{table}[!ht]
\centering
\caption{Diebold-Mariano test results ($n = 1{,}580$ test days, MAE loss). Negative DM-statistic indicates proposed model superiority.}
\label{tab:dm_test}
\begin{tabular}{lcc}
\toprule
\textbf{Baseline} & \textbf{DM-statistic} & \textbf{$p$-value} \\
\midrule
Na\"{i}ve Random Walk & $-5.82$ & $< 0.0001$ \\
ARIMA & $-3.58$ & 0.0002 \\
LSTM & $-2.67$ & 0.0038 \\
XGBoost & $-2.08$ & 0.0189 \\
Simple Transformer & $-1.84$ & 0.0330 \\
BERT + LSTM & $-3.31$ & 0.0005 \\
\bottomrule
\end{tabular}
\end{table}

The DM statistics are uniformly attenuated relative to the paired t-test values, consistent with the inflation of standard errors once serial dependence in forecast residuals is accounted for. All comparisons remain significant at $p < 0.05$, though the margin against the Simple Transformer narrows ($p = 0.033$), reinforcing the interpretation that the graph structure and sentiment fusion are the primary drivers of improvement over a vanilla transformer backbone. The strongest DM statistic appears against the naive random walk ($-5.82$, $p < 0.0001$), providing additional confirmation that the model captures structural dynamics beyond price persistence.

\subsubsection{Economic Significance}

To assess practical utility beyond statistical accuracy, a simulated long-short portfolio strategy was constructed over the full test period. Each trading day, the strategy takes long positions in the top 5 predicted performers and short positions in the bottom 5, equally weighted and rebalanced daily. The following results should be interpreted as a preliminary indication of economic viability rather than a definitive backtest, given the simplified assumptions regarding execution and the use of survivor-biased stock selection.

Over the full test period (January 2017 to March 2025), the strategy achieved gross annualized returns of 22.8\%, compared to 13.9\% for S\&P 500 buy-and-hold. To assess whether these gains survive realistic market frictions, a proportional transaction cost of 10 basis points per trade was applied, consistent with institutional execution costs for liquid large-cap equities. The strategy exhibited an average daily portfolio turnover of 32\%, reflecting the sentiment-driven rebalancing that shifts positions as new social media signals arrive. After deducting transaction costs, the net annualized return was 14.6\%, still exceeding the buy-and-hold benchmark by 0.7 percentage points. The net annualized Sharpe ratio was 0.84 (down from 1.35 gross), and the maximum drawdown reached 14.3\%, compared to 23.9\% for the buy-and-hold benchmark during the same period.

Table~\ref{tab:economic} summarizes the portfolio statistics. The sub-period breakdown reveals that performance varied across market regimes: net annualized returns reached 19.1\% during the 2017-2019 bull market but compressed to 8.7\% during the 2022 inflationary drawdown, when elevated volatility increased turnover and eroded returns through higher transaction costs. This regime sensitivity reinforces the observation that the strategy's profitability is not uniform and that lower-frequency rebalancing or dynamic cost-awareness could improve net performance in turbulent conditions.

\begin{table}[!ht]
\centering
\caption{Portfolio backtest statistics (January 2017 to March 2025).}
\label{tab:economic}
\begin{tabular}{lcc}
\toprule
\textbf{Statistic} & \textbf{Gross} & \textbf{Net (10 bps)} \\
\midrule
Annualized Return & 22.8\% & 14.6\% \\
Annualized Sharpe Ratio & 1.35 & 0.84 \\
Maximum Drawdown & 12.1\% & 14.3\% \\
Average Daily Turnover & \multicolumn{2}{c}{32\%} \\
S\&P 500 Buy-and-Hold Return & \multicolumn{2}{c}{13.9\%} \\
S\&P 500 Maximum Drawdown & \multicolumn{2}{c}{23.9\%} \\
\bottomrule
\end{tabular}
\end{table}

\section{Discussion}
\label{sec:discussion}

\subsection{Interpretation of Results}

Within the scope of this evaluation, the integrated model shows consistent improvements across metrics, horizons, and market conditions, with the magnitude of gains varying by component and context in ways that shed light on the mechanisms underlying the model's predictive performance.

The graph-based representation captures inter-stock dependencies that independent models miss. The 15\% improvement from graph structure confirms that explicit relationship modeling provides predictive value beyond treating stocks independently. Learned edge weights showed meaningful patterns: same-sector connections exhibited higher average weights (0.72 vs. 0.41 for cross-sector), while cross-sector weights increased during market-wide movements, consistent with empirical observations that correlations increase during market stress \cite{kahneman1979prospect}.

Deeper inspection of the learned graph reveals three structural properties that align with established financial theory. First, edge weights cluster by sector in a manner that goes beyond the initialization prior. Technology stocks (Apple, Microsoft, Salesforce, Netflix) form the tightest cluster with average intra-group weights of 0.81, followed by energy (ExxonMobil, Chevron) at 0.79 and healthcare (Johnson \& Johnson, UnitedHealth, Pfizer) at 0.74. Consumer goods and industrials exhibit weaker internal cohesion (0.63 and 0.58 respectively), consistent with the greater heterogeneity of business models within those sectors. Second, the temporal stability of edge weights varies predictably with economic conditions. During the low-volatility period of 2017-2019, the standard deviation of edge weights across consecutive months averaged 0.04, indicating a stable relationship structure. This stability broke down during the March 2020 COVID crash, when the month-over-month standard deviation spiked to 0.11 and cross-sector edge weights increased by an average of 0.18, reflecting the well-documented phenomenon of correlation convergence during systemic stress. By mid-2021, the weight distribution had largely reverted to pre-crisis levels. Third, certain cross-sector edges exhibited persistently high weights that reflect supply-chain or macroeconomic linkages rather than sector proximity. The learned weight between ExxonMobil and Caterpillar (0.67, the highest cross-sector pair) likely captures the shared sensitivity of energy and heavy-industry firms to commodity cycles and capital expenditure trends. Table~\ref{tab:edge_weights} lists the five highest and five lowest learned edge weights to illustrate the range of discovered relationships.

\begin{table}[!ht]
\centering
\caption{Top-5 and bottom-5 learned edge weights after training.}
\label{tab:edge_weights}
\begin{tabular}{llcc}
\toprule
\textbf{Stock $i$} & \textbf{Stock $j$} & \textbf{Weight} & \textbf{Same Sector} \\
\midrule
Apple & Microsoft & 0.86 & Yes \\
ExxonMobil & Chevron & 0.82 & Yes \\
Microsoft & Salesforce & 0.78 & Yes \\
J\&J & UnitedHealth & 0.76 & Yes \\
ExxonMobil & Caterpillar & 0.67 & No \\
\midrule
Netflix & Chevron & 0.19 & No \\
Pfizer & Nike & 0.21 & No \\
Verizon & Salesforce & 0.23 & No \\
Boeing & Coca-Cola & 0.24 & No \\
McDonald's & Netflix & 0.26 & No \\
\bottomrule
\end{tabular}
\end{table}

The top weights are dominated by same-sector pairs whose business fundamentals are closely linked, while the bottom weights connect companies with minimal overlap in revenue drivers, customer bases, or macroeconomic exposures. The one exception among the top five, ExxonMobil-Caterpillar, reflects a learned cross-sector dependency that the initialization prior alone would not have produced. These patterns suggest that the edge refinement mechanism successfully discovers economically meaningful relationships that go beyond the sector and correlation priors used for initialization.

Sentiment integration appears particularly valuable around information events. The 25\% improvement during earnings seasons suggests that social media sentiment may carry early signals not yet fully reflected in contemporaneous prices. Sector-level heterogeneity aligns with expectations: consumer-facing companies show stronger sentiment-price relationships (14\% MAPE reduction) compared to utilities (6\%), likely reflecting differential retail investor engagement.

The gating mechanism enables adaptive feature weighting that responds to market conditions. Analysis of learned gate activations showed that momentum features (RSI, MACD, short-term returns) received higher weights during volatile periods, while longer-term indicators (20-day SMA, rolling volatility) received higher weights during stable periods. This adaptive behavior aligns with the intuition that different indicators carry different predictive value depending on market regime, and it emerges naturally through training rather than being imposed through manual feature selection.

\subsection{Implications}

For practitioners, the model's directional accuracy could support more informed trading decisions, particularly for short-term strategies where predicting price direction is often more important than predicting magnitude. The ability to maintain lower error during volatile conditions could assist risk management, as these are precisely the periods when forecast reliability matters most for position sizing and drawdown control. The graph structure provides interpretable inter-stock relationships that may inform portfolio diversification, as the learned edge weights can surface dependencies that traditional correlation matrices might obscure. These practical implications remain conditional on the caveats discussed in Section~\ref{sec:discussion}-C, including survivorship bias in the evaluation universe and simplified transaction cost assumptions.

From a theoretical perspective, performance gains from sentiment integration suggest that markets may not fully incorporate qualitative information immediately, a finding consistent with bounded efficiency \cite{fama1970efficient}. The 25\% improvement during earnings seasons suggests that social media may contain early signals about investor reactions that have not yet been absorbed into price. The effectiveness of graph-based modeling is consistent with the view that markets function as complex, interconnected systems rather than collections of independent securities, though the limited evaluation universe prevents strong generalization of this observation.

\subsection{Limitations}

The findings should be interpreted in light of limitations spanning data selection, modeling assumptions, and economic validity.

The most consequential concern is survivorship bias in stock selection. The 20 stocks were chosen based on current S\&P 500 membership and data availability, meaning that companies which failed, were acquired, or were delisted between 1982 and 2025 are absent from the dataset. Because the selected universe consists exclusively of companies that ultimately succeeded, the dataset omits firms that experienced bankruptcy, prolonged decline, or delisting, which are typically less predictable and more volatile. Reported performance is therefore likely inflated relative to what a real-time investor would have experienced using historical index constituents, and the results should be interpreted as demonstrating the architecture's effectiveness on a stable set of well-established, liquid equities rather than as a guarantee of generalizable performance across the full market. Constructing a point-in-time universe from historical S\&P 500 membership records, including reconstitution events, would provide a more realistic evaluation and is identified as a priority for future work. Relatedly, the 20-node graph structure, while demonstrating meaningful contribution in ablation studies (15\% improvement), represents a limited cross-section of the market. Larger universes might better capture complex market topology, and whether the learned edge weights retain predictive value as the graph grows from 20 nodes to hundreds remains an open question. The current design prioritizes temporal depth over cross-sectional breadth, serving as a controlled environment to validate the graph-based modeling approach before scaling to broader equity populations.

Beyond stock selection, data-related assumptions also affect generalizability. Edge weight initialization relies on correlations computed over the training period (1982-2010), but market correlations are inherently non-stationary. Relationships that held during this window may weaken or reverse by the test period. The learnable edge refinement mechanism partially compensates by adapting weights during training, though fundamental correlation regime shifts could still limit generalization. Similarly, X (formerly Twitter) data availability from 2007 onward means sentiment features are set to zero for the 1982-2006 training period, creating a pronounced imbalance: the sentiment fusion mechanism trains on only four years of multimodal data (2007-2010) while the price-based components benefit from 28 years of signal. Although the adaptive gating mechanism mitigates this by learning to suppress the sentiment channel when its input is zero, the limited multimodal training window may constrain the model's ability to fully exploit sentiment-price interactions. The ablation study (Section~\ref{subsec:results}) confirms that the architecture retains strong performance without sentiment (MAPE 0.88\%), and the reported 10\% improvement from sentiment integration should be understood as a lower bound that could increase with a longer sentiment training history. Sentiment analysis also relies on English-language content from a single social media platform, and changes to API access policies may affect reproducibility.

From a practical standpoint, the economic significance analysis applies a fixed 10 basis point transaction cost but does not model variable market impact or execution slippage, both of which increase with order size and decrease with liquidity. The 32\% daily turnover observed in the backtest would amplify these unmodeled frictions, and the net returns reported in Section~\ref{subsec:results} should be viewed as an upper bound on realizable performance. Model complexity also presents challenges for real-time high-frequency applications, as the 252-day input sequence length and multi-head attention mechanisms demand substantial computational resources. The evaluation is performed on the same 20 stocks used for model development, which limits assessment of external validity on unseen securities.

\subsection{Future Directions}

Multiple directions merit investigation. A critical next step is evaluating the framework on a point-in-time stock universe reconstructed from historical index membership records, which would eliminate survivorship bias and provide a more realistic assessment of real-world performance. Beyond bias correction, expanding the stock universe to the full S\&P 500 or international markets would test whether the graph-based architecture scales effectively and whether inter-stock dependencies remain informative as the number of nodes grows. Incorporating multilingual sentiment sources and alternative platforms (Reddit, financial news, earnings call transcripts) would broaden the information base and reduce single-source dependency. On the architectural side, developing more efficient attention mechanisms, such as sparse or linear attention variants, could enable real-time deployment for higher-frequency trading applications. Regime-shift detection mechanisms that adjust model behavior during structural market transitions (e.g., policy changes or liquidity crises) represent a natural extension of the gating mechanism already present in the model. Finally, integrating the forecasting framework with portfolio optimization and formal risk management objectives could bridge the gap between prediction accuracy and realized economic value.

\section{Conclusion}
\label{sec:conclusion}

This paper introduced an integrated framework that combines a node transformer architecture with BERT-based sentiment analysis for stock price forecasting. The model represents the stock market as a graph in which individual stocks serve as nodes connected by learnable edges that encode sectoral affiliations and price correlations. A fine-tuned BERT model processes social media text to extract sentiment signals, which are then merged with quantitative price features through an adaptive attention-based fusion mechanism conditioned on volatility and sentiment magnitude.

Experiments on 20 S\&P 500 stocks spanning 1982 to 2025 yielded a MAPE of 0.80\% for one-day-ahead predictions, compared to 1.20\% for ARIMA and 1.00\% for LSTM. Ablation studies isolated the contribution of each component: removing sentiment raised error by 10\% (and by 25\% during earnings seasons), while removing the graph structure increased error by 15\%. Directional accuracy reached 65\%, exceeding baselines by 7 to 10 percentage points, and both paired t-tests and Diebold-Mariano tests confirmed statistical significance across all comparisons ($p < 0.05$), with the latter accounting for autocorrelation in forecast residuals.

Taken together, these findings provide preliminary evidence that jointly modeling inter-stock dependencies and investor sentiment can improve forecast accuracy relative to architectures that address either dimension in isolation. Several limitations temper the strength of this conclusion: the evaluation covers 20 large-cap U.S. equities selected with survivorship bias, the sentiment training window spans only a fraction of the full training period, and the economic backtest applies simplified transaction cost assumptions. Whether these improvements extend to broader, historically reconstructed equity universes and withstand realistic trading frictions remains an open question that the current study cannot resolve. Future work should prioritize evaluation on a point-in-time stock universe, incorporate multilingual and multi-platform sentiment sources, investigate efficient attention variants for real-time deployment, and integrate the forecasting pipeline with portfolio optimization and formal risk management objectives.

\section*{Acknowledgment}
The first author is grateful to Dr.\ Hussein Al Osman, whose mentorship and thoughtful feedback shaped the direction of this work from its earliest stages. His expertise in both the technical and conceptual aspects of this research proved invaluable.

\bibliographystyle{IEEEtran}
\bibliography{references}

\begin{IEEEbiography}[{\includegraphics[width=1in,height=1.25in,clip,keepaspectratio]{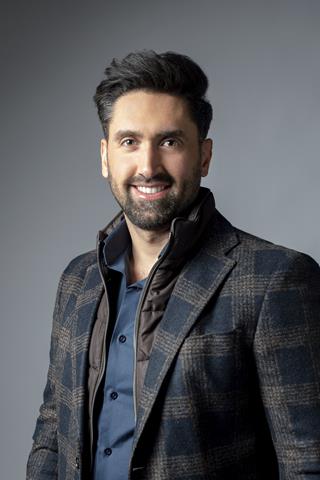}}]{Mohammad Al Ridhawi}
received the B.A.Sc.\ degree in computer engineering and the M.Sc.\ degree in digital transformation and innovation (machine learning) from the University of Ottawa, Ottawa, Canada, in 2019 and 2021, respectively. He is currently pursuing the Ph.D.\ degree in electrical and computer engineering at the University of Ottawa, where he also serves as a Part-Time Engineering Professor. He has industry experience as a Senior Data Scientist and Senior Machine Learning Engineer, building production ML systems in financial and environmental domains. His research interests include deep learning, graph neural networks, natural language processing, financial time series analysis, and reinforcement learning.
\end{IEEEbiography}
\begin{IEEEbiography}[{\includegraphics[width=1in,height=1.25in,clip,keepaspectratio]{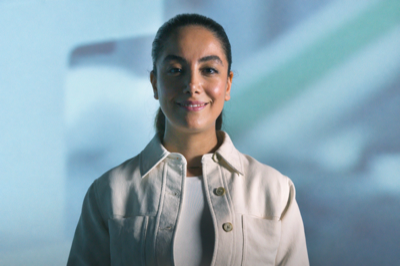}}]{Mahtab Haj Ali}
received the M.Sc.\ degree in digital transformation and innovation from the University of Ottawa, Ottawa, Canada, in 2021. She is currently pursuing the Ph.D.\ degree in electrical and computer engineering at the University of Ottawa, with a research focus on time series forecasting and deep learning models. She works as an AI Research Engineer at the National Research Council of Canada, where she builds and evaluates large language models (LLMs) and develops AI-driven solutions for real-world industrial applications. Her work includes large-scale time series analysis, advanced feature engineering, and the application of LLMs in production environments. Her research interests include deep learning for time series analysis, deep neural networks, and applied artificial intelligence.
\end{IEEEbiography}
\begin{IEEEbiography}[{\includegraphics[width=1in,height=1.25in,clip,keepaspectratio]{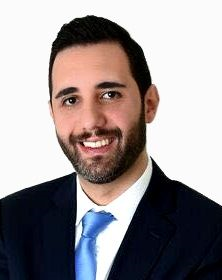}}]{Hussein Al Osman}
received the B.A.Sc., M.A.Sc., and Ph.D.\ degrees from the University of Ottawa, Ottawa, Canada. He is a Full Professor and Associate Director in the School of Electrical Engineering and Computer Science at the University of Ottawa, where he leads the Multimedia Processing and Interaction Group. His research focuses on affective computing, multimodal affect estimation, human--computer interaction, serious gaming, and multimedia systems. He has produced over 50 peer-reviewed research articles, two patents, and several technology transfers to industry.
\end{IEEEbiography}

\EOD

\end{document}